\documentclass{article}
\pdfoutput=1

    \PassOptionsToPackage{numbers, compress}{natbib}
\usepackage[final]{neurips_2022}

\usepackage[utf8]{inputenc} % allow utf-8 input
\usepackage[T1]{fontenc}    % use 8-bit T1 fonts
\usepackage{hyperref}       % hyperlinks
\usepackage{url}            % simple URL typesetting
\usepackage{booktabs}       % professional-quality tables
\usepackage{amsfonts}       % blackboard math symbols
\usepackage{nicefrac}       % compact symbols for 1/2, etc.
\usepackage{microtype}      % microtypography
\usepackage{xcolor}         % colors

\usepackage{amsmath}
\usepackage{caption}
\usepackage{subcaption}
\usepackage{algorithm}
\usepackage{algorithmic}
\usepackage{wrapfig}

\newcommand{\bfx}{\mathbf{x}}
\newcommand{\bfy}{\mathbf{y}}

\newcommand{\bfK}{\mathbf{K}}

\newcommand{\bfW}{\mathbf{W}}
\newcommand{\bfT}{\mathbf{T}}
\newcommand{\bfH}{\mathbf{H}}

\definecolor{lineBlue}{RGB}{57,106,177}
\definecolor{lineOrange}{RGB}{218,124,48}
\definecolor{lineGreen}{RGB}{62,150,81}
\definecolor{lineRed}{RGB}{204,37,41}
\definecolor{lineGray}{RGB}{83,81,84}
\definecolor{linePurple}{RGB}{107,76,154}
\definecolor{lineMaroon}{RGB}{146,36,40}
\definecolor{barBlue}{RGB}{114,147,203}
\definecolor{barOrange}{RGB}{225,151,76}
\definecolor{barGreen}{RGB}{132,186,91}
\definecolor{barRed}{RGB}{211,94,96}
\definecolor{barGray}{RGB}{128,133,133}
\definecolor{barPurple}{RGB}{144,103,167}
\definecolor{barMaroon}{RGB}{171,104,81}

\usepackage{pgfplots}
\pgfplotsset{compat=1.17}

\title{Wavelet Feature Maps Compression \\
           for Image-to-Image CNNs}

\author{%
  Shahaf E.~Finder
  \thanks{Contributed equally.} \ ,
  Yair Zohav $^*$,
  Maor Ashkenazi $^*$,
  Eran Treister
 \\
  The Department of Computer Science, Ben-Gurion University \\
  {\small \tt [finders,maorash]@post.bgu.ac.il \, erant@cs.bgu.ac.il }
}

\begin{document}

\maketitle

\begin{abstract}
Convolutional Neural Networks (CNNs) are known for requiring extensive computational resources, and quantization is among the best and most common methods for compressing them.
While aggressive quantization (i.e., less than 4-bits) performs well for classification, it may cause severe performance degradation in image-to-image tasks such as semantic segmentation and depth estimation.
In this paper, we propose Wavelet Compressed Convolution (WCC)---a novel approach for high-resolution activation maps compression integrated with point-wise convolutions, which are the main computational cost of modern architectures. 
To this end, we use an efficient and hardware-friendly Haar-wavelet transform, known for its effectiveness in image compression, and define the convolution on the compressed activation map. We experiment with various tasks that benefit from high-resolution input. By combining WCC with light quantization, we achieve compression rates equivalent to 1-4bit activation quantization with relatively small and much more graceful degradation in performance. 
Our code is available at \url{https://github.com/BGUCompSci/WaveletCompressedConvolution}.
\end{abstract}

\section{Introduction} \label{sec:intro}

Over the past years, Convolutional Neural Networks (CNNs) have brought significant improvement in processing images, video, and audio \cite{lecun2015deep,krizhevsky2017imagenet}. 
However, CNNs require significant computational and memory costs, which makes the usage of CNNs difficult in applications where computing power is limited, \textit{e.g.},~on edge devices. To address this limitation, several approaches have been proposed to reduce the computational costs of neural networks.
Among the popular ones are weight pruning \cite{han2015learning, guo2016dynamic}, architecture search \cite{howard2019searching,kirillov2020pointrend}, and quantization \cite{li2017trainingquantnets, banner2018scalable}. In principle, all these approaches can be applied simultaneously on top of each other to reduce the computational costs of CNNs.

Specifically, the quantization approach relieves the computational cost of CNNs by quantizing their weights and activation (feature) maps using low numerical precision so that they can be stored and applied as fixed point integers \cite{hubara2017quantized, banner2018scalable}.
In particular, it is common to apply aggressive quantization (less than 4-bit precision) to compress the activation maps \cite{esser2019lsq}. However, it is known that compressing natural images using uniform quantization is sub-optimal. Indeed, applying aggressive quantization in certain CNNs can lead to significant degradation in the network's performance. The impact is especially evident for image-to-image tasks such as semantic segmentation \cite{tang2019towards} and depth prediction \cite{godard2019digging}, where each pixel has to be assigned a value. \textit{E.g.},~in a recent work that targets quantized U-Nets \cite{tang2019towards}, activations bit rates are kept relatively high ($\sim 8$ bits) while the weight bit rates are lower (down to 2 bits). Beyond that, we note that the majority of the quantization works are applied and tested on image classification \cite[and references therein]{esser2019lsq, li2019additive} and rarely on other tasks.

This work aims to improve the compression of the activation maps by introducing Wavelet Compressed Convolution (WCC) layers, which use wavelet transforms to compress activation maps before applying convolutions. To this extent, we utilize Haar-wavelet transform \cite{daubechies1992ten} due to our ability to apply it (and its inverse) efficiently in linear complexity for each channel, using additions and subtractions only, thanks to the simplicity of the Haar basis. The core idea of our approach is to keep the same top $k$ entries in magnitude of the transformed activation maps with respect to \emph{all channels} (dubbed as joint shrinkage) and perform the convolution in the wavelet domain on the \emph{compressed} signals, saving significant computations. We show that the transform and shrinkage operations commute with the $1\times1$ (point-wise) convolution, the heart of modern CNNs. This procedure is applied along with modest quantization to reduce computational costs further.

To summarize, our key contributions are:
1) We propose Wavelet Compressed Convolution (WCC), a novel approach to compress $1\times1$ convolutions using a modified wavelet compression technique.
2) We demonstrate that applying low-bit quantization on popular image-to-image CNNs may yield degradation in performance. Specifically, we show that for object detection, semantic segmentation, depth prediction, and super-resolution.
3) We show that using WCC dramatically improves the results for the same compression rates, using it as a drop-in replacement for the $1\times 1$ convolutions in the baseline network architectures.

%\begin{wrapfigure}{R}{0.5\textwidth}
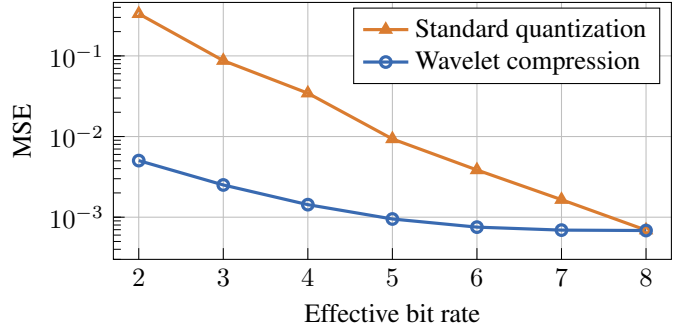
\begin{figure}
\centering
\begin{tikzpicture}
% \definecolor{color0}{rgb}{0.12156862745098,0.466666666666667,0.705882352941177}
% \definecolor{color1}{rgb}{1,0.498039215686275,0.0549019607843137}
\begin{axis}[
legend cell align={left},
width=9cm,height=5cm,
xmajorgrids=true,
ymajorgrids=true,
log basis y={10},
tick pos=left,
% x grid style={white!69.0196078431373!black},
xlabel={Effective bit rate},
xmin=1.7, xmax=8.3,
xtick style={color=black},
% y grid style={white!69.0196078431373!black},
ylabel={MSE},
ymin=0.0003, ymax=0.46,
ymode=log,
ytick style={color=black}
]
\addplot [draw=lineOrange, mark=triangle, very thick]
table{%
x  y
2 0.332652598619461
3 0.0873541310429573
4 0.0344834066927433
5 0.00933729019016027
6 0.00385086005553603
7 0.00165022211149335
8 0.000691189663484693
};
\addlegendentry{Standard quantization}
\addplot [draw=lineBlue, mark=o, very thick]
table{%
x  y
2 0.00503227300941944
3 0.00251046125777066
4 0.00143062637653202
5 0.000950176268815994
6 0.000753191998228431
7 0.000691640074364841
8 0.000683698803186417
};
\addlegendentry{Wavelet compression}
\end{axis}
\end{tikzpicture}
\caption{Comparison of standard quantization and our proposed wavelet compression. The ordinate represents MSE between the quantized and original activation maps based on $10^3$ activation maps from the second hidden layer of MobileNetV3 (small) using the ImageNet data set.}
\label{fig:mse_compare}
\end{figure}
%\end{wrapfigure}

\section{Related Work} \label{sec:related}
\textbf{Quantized neural networks.}
Quantized neural networks have been quite popular recently and are exhibiting impressive progress in the goal of true network compression and efficient CNN deployment. Quantization methods include \cite{zhou2018dorefanet,zhang2018lqnets,Banner2018ACIQAC}, and in particular \cite{li2019additive,esser2019lsq,choi2019accurate}, which show that the clipping parameters---an essential parameter in quantization schemes---can be learned through optimization. Beyond that, there are more sophisticated methods to improve the mentioned schemes. For example, dynamic quantization schemes utilize different bit allocations at every layer \cite{dong2019hawqv2,cai2020rethinking,mpd}. Non-uniform methods can improve the quantization accuracy \cite{yamamoto2021learnable} but require a look-up table, which reduces hardware efficiency. Quantization methods can also be enhanced by combination with pruning \cite{tung2018deep} and knowledge distillation for better training \cite{kim2019qkd}. 

The works above focus on image classification. When considering image-to-image tasks (\textit{e.g.}~semantic segmentation) networks tend to be more sensitive to quantization of the activations. In the work of \cite{askarihemmat2019u}, targeting segmentation of medical images, the lowest bit rate for the activations is 4 bits, and significant degradation in the performance is evident compared to 6 bits. These results are consistent with the work of \cite{tang2019towards} that was mentioned earlier, which uses a higher bit rate for the activations than for the weights. \cite{xu2018quantization} use weight (only) quantization for medical image segmentation as an attempt to remove noise and not for computational efficiency. The recent work of \cite{liu2021zero} shows both a sophisticated post-training quantization scheme and includes fine-tuned semantic segmentation results. Again a significant degradation is observed when going from 6 to 4 bits. One exception is the work of \cite{heinrich2018ternarynet} that uses ternary networks (equivalent to 2 bits here) and segments one medical data set relatively well compared to its full-precision baseline.

In this work, we focus on the simplest possible quantization scheme: uniform (across all weights and activations), quantization-aware training, and per-layer clipping parameters. That is to ensure hardware compatibility and efficient use of available training data. In principle, one can run our platform regardless of the quantization type and scenario (\textit{e.g.},~non-uniform/mixed precision quantization). Also, since we target activations' compression, any method focused on the weights (\textit{e.g.},~pruning \cite{tung2018deep}) can also be combined with our proposed method. 

\textbf{Wavelet transforms in neural networks.}
Wavelet transforms are widely used in image processing \cite{porwik2004haar}. For example, the JPEG2000 format \cite{rabbani2002jpeg2000} uses the wavelet domain to represent images as highly sparse feature maps. Recently, wavelet transforms have been used to define convolutions and architectures in CNNs for various imaging tasks: \cite{gal2021swagan} incorporate wavelets in Generative Adversarial Networks to enhance the visual quality of generated images as well as improve computational performance; \cite{huang2017wavelet} present a network architecture for super-resolution, where the wavelet coefficients are predicted and used to reconstruct the high-resolution image; \cite{duan2017sar} and \cite{williams2018wavelet} use wavelet transforms in place of pooling operators to improve CNNs performance. The former uses a dual-tree complex wavelet transform \cite{kingsbury1998dual}, and the latter learns the wavelet basis as part of the network optimization. In all of these cases, the wavelet transform is not used for compression but rather to preserve information with its low-pass filters.
\cite{liu2018multi} suggest using a modified U-Net architecture for image-to-image translation. There, wavelet transforms are used for down-sampling, and the inverse is used for up-sampling. This work is architecture-specific, and the method can not be easily integrated into other architectures. In contrast, our proposed WCC layer can easily replace $1\times 1$ convolutions regardless of the CNN architecture. Hence our framework is, in principle, also suitable for post-training quantization \cite{soudry1}, where the data is unavailable, and the original network is not retrained.

In the context of compression, \cite{wolter2020neural} propose a wavelet-based approach to learn basis functions for the wavelet transform to compress the weights of linear layers, as opposed to compression of the activation as we apply here. We use the Haar transform (as opposed to a learned one) for its hardware efficiency. Using a different transform (known or learned) is also possible at the corresponding computational cost. \cite{sun2021mwq} introduce quantization in the wavelet domain, as we do in this work. However, the authors suggest improving the quantization scheme by learning a different clipping parameter per wavelet component, but without the feature shrinkage stage, which is the heart of our approach (we use the same clipping parameter for the whole layer using 8 bits for hardware efficiency). As mentioned before, one can use our WCC together with different types of quantization schemes, including the one proposed by \cite{sun2021mwq}, taking into account the additional hardware complexity of using different clipping parameters per component.

\section{Background}
\label{sec: quant-aware-train}

\textbf{Quantization-aware training.} Quantization schemes can be divided into post- and quantization-aware training schemes. Post-training schemes perform model training and model quantization separately, which is most suitable when the training data is unavailable during the quantization phase \cite{soudry1, DFQ,cai2020zeroq}.  
On the other hand, quantization-aware training schemes are used to adapt the model's weights as an additional training phase. Such schemes do require training data but generally provide better performance. Quantization schemes can also be divided into uniform vs. non-uniform methods, where the latter is more accurate, but the former is more hardware-friendly \cite{li2019additive, jung2019learning}. Lastly, quantization schemes can utilize quantization parameters per-channel within each layer or utilize these parameters only per layer (where all the channels share the same parameters). Similar to before, per-channel methods are difficult to exploit in hardware, while per-layer methods are less accurate but more feasible for deployment on edge devices. This paper focuses on per-layer and uniform quantization-aware training for both weights and activations and aims to improve on top of it. Other quantization schemes can be applied within our wavelet compression framework as well. 

In quantization-aware training, we set the values of the weights to belong to a small set so that after training, the application of the network can be carried out in integer arithmetic, \textit{i.e.},~activation maps are quantized as well. Even though we use discontinuous rounding functions throughout the network, quantization-aware training schemes utilize gradient-based methods to optimize the network's weights \cite{han2015learning, yin2019understanding}. In a nutshell,
when training, we iterate on the floating-point values of the weights. During the forward pass, both the weights and activation maps are quantized, while during the backward pass, the Straight Through Estimator (STE) approach is used \cite{bengio2013estimating}, where we ignore the rounding function, whose exact derivative is zero.

The specific quantization scheme that we use is based on the work of \cite{li2019additive}. First, the pointwise quantization operator is defined by:
\begin{equation} \label{eq:quantoperator}
 q_b(t) = \textstyle{\frac{\mbox{round}((2^b - 1) \cdot t)}{2^b - 1}},
\end{equation}
where $t\in[-1,1]$ or $t\in[0,1]$ for signed or unsigned quantization, respectively\footnote{In most standard CNNs, the ReLU activation is used; hence the activation feature maps are non-negative and can be quantized using an unsigned scheme. If a different activation function that is not non-negative is used, or, as in our case, the wavelet coefficients are quantized, signed quantization should be used instead.}. The parameter $b$ is the number of bits that are used to represent $t$. Each entry undergoes three steps during the forward pass: scale, clip, and round. That is, to get the quantized version of a number, we apply: 
\begin{equation}\label{eq:quantweights}
x_b = 
\begin{cases} 
      \alpha q_{b-1}(\mbox{clip}(\frac{x}{\alpha}, -1, 1)) & \mbox{if signed} \\
      %Q_b(x) = 
      \alpha q_{b}(\mbox{clip}(\frac{x}{\alpha}, 0, 1)) & \mbox{if unsigned} \\
  \end{cases}.
\end{equation}
Here, $x, x_b$ are the real-valued and quantized tensors, and $\alpha$ is the clipping parameter. 
The parameters $\alpha$ in \eqref{eq:quantweights} play an important role in the error generated at each quantization and should be chosen carefully. Recently, \cite{li2019additive,esser2019lsq} introduced an effective gradient-based optimization to find the clipping values $\alpha$ for each layer, again using the STE approximation. This enables the quantized network to be trained in an end-to-end manner with backpropagation.
To further improve the optimization, weight normalization is also used before each quantization.

\textbf{Haar wavelet transform and compression.} In this section, we describe the Haar-wavelet transform in deep learning language and its usage for compression. See \cite{vyas2018multiscale} for more details on the use of wavelets for image compression. Given an image channel $\bfx$, the one-level Haar transform can be achieved by a separable 2D convolution with a stride of 2 using the following weight tensor:
\begin{eqnarray}\label{eq:HaarKernel}
\bfW = \frac{1}{2}\begin{bmatrix}
\begin{bmatrix}
\;1 & \;\;1 \\
\;1 & \;\;1
\end{bmatrix}
,
\begin{bmatrix}
\;1 & -1 \\
\;1 & -1
\end{bmatrix}
,
\begin{bmatrix}
\;\;1 & \;\;1 \\
-1 & -1
\end{bmatrix},
\begin{bmatrix}
\;\;1 & -1 \\
-1 & \;\;1
\end{bmatrix}
\end{bmatrix}.
\end{eqnarray}
These kernels can also be expressed as a composition of separable 1D kernels $[1,1]/\sqrt{2}$ and $[1,-1]/\sqrt{2}$.
The result of the convolution $\left[\bfy_1,\bfy_2,\bfy_3,\bfy_4\right] = \mbox{Conv}(\bfW,\bfx)$ has four channels, each of which has half the resolution of $\bfx$. The leftmost kernel in \eqref{eq:HaarKernel} is an averaging kernel, and the three right kernels are edge-detectors. This, together with the fact that images are piece-wise smooth, leads to relatively sparse images $\bfy_2,\bfy_3,\bfy_4$. Hence, if we retain only the few top-magnitude entries in these vectors, we keep most of the information, as most of the entries we drop are zeros. That is the main idea of wavelet compression. We denote the Haar-wavelet transform by $\bfy = \mbox{HWT}(\bfx)$, where $\bfy$ is defined as the concatenation of the vectors $\bfy_1,...,\bfy_4$ into one. Since the kernels in \eqref{eq:HaarKernel} form an orthonormal basis, applying the inverse transform is obtained by the transposed convolution of \eqref{eq:HaarKernel}: 
\begin{align}
\bfx = \mbox{iHWT}(\bfy) = \mbox{Conv-transposed}(\bfW,\left[\bfy_1,\bfy_2,\bfy_3,\bfy_4\right]). 
\end{align}
However, unlike $\bfy_2,\bfy_3,\bfy_4$, the image $\bfy_1$ is not sparse, and is just the down-sampled $\bfx$. Hence, in the multilevel wavelet transform, we recursively apply the convolution with $\bfW$ on low-pass filtered sub-bands $\bfy_1$ to generate further down-sampled sparse channels. For example, a 2-level Haar transform can be summarized as
\begin{equation}
\mbox{Conv}(\bfW,\bfx) = \left[\bfy_1^{1},\bfy_2^{1},\bfy_3^{1},\bfy_4^{1}\right]; \quad
\mbox{Conv}(\bfW,\bfy_1^1) = \left[\bfy_1^{2},\bfy_2^{2},\bfy_3^{2},\bfy_4^{2}\right],
% \end{aligned}
\end{equation}

and the resulting transformed image with 2-levels can be written as the concatenated vector
\begin{equation}\label{eq:wavelet_transform}
\begin{aligned}
\mbox{HWT}(\bfx)=\bfy = \left[\bfy_1^{2},\bfy_2^{2},\bfy_3^{2},\bfy_4^{2},\bfy_2^{1},\bfy_3^{1},\bfy_4^{1}\right].
\end{aligned}
\end{equation}
In this work we use 3 levels in all the experiments. To apply compression we define the operator $\bfT$ that extracts the top $k$ entries in magnitude from the vector $\bfy$: 
\begin{equation}\label{eq:T}
\bfy^{compressed} = \bfT\cdot \mbox{HWT}(\bfx).
\end{equation} To de-compress this vector we first zero-fill $\bfy^{compressed}$ (\textit{i.e.},~multiply with $\bfT^\top$) and apply the inverse Haar transform, which involves with the transposed operations in the opposite order. 

\section{Wavelet Compressed Convolution} \label{sec:WaveletMethod}

\usetikzlibrary{3d,decorations.text,shapes.arrows,positioning,fit,backgrounds}

\begin{figure*}
\centering
\resizebox{\textwidth}{!}{
\begin{tikzpicture}[x={(1,0)},y={(0,1)},z={({cos(40)},{sin(40)})},
font=\sffamily\small,scale=2.5]

\foreach \X [count=\Z]in {3,2,1}
 {\node[opacity=1.] at (\Z/4,0,1-\Z/4) {\includegraphics[width=2.5cm]{images/im\X.png}};}

\draw [-to](1.4,0.5) -- (1.7,0.5) node [pos=0.5, above, font=\footnotesize] {A};

\foreach \X [count=\Z]in {3,2,1}
 {\node[opacity=1.] at (1.4+\Z/4,0,1-\Z/4) {\includegraphics[width=2.5cm]{images/im\X_wt.png}};}

\draw [-](2.8,0.5) -- (3.1,0.5) node [pos=0.5, above, font=\footnotesize] {B};

\draw [-](3.1,0.5) -- (3.1,0);
\draw [-to](3.1,0) -- (3.3,0);

\node[opacity=1.] at (3.45,-0.35,0.5) {\includegraphics[width=2.5cm]{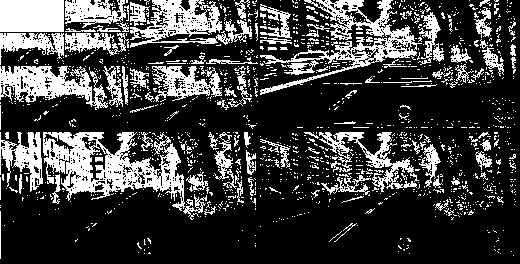}};

\draw [-](4.35,0) -- (4.55,0);
\draw [-](4.55,0) -- (4.55,0.5);

\draw [-](3.1,0.5) -- (3.1,1);
\draw [-to](3.1,1) -- (3.35,1);

\foreach \X [count=\Z]in {a,b,c}
 {\draw[draw=black, fill=black!20] (2.6+\Z/8, -0.15, 1-\Z/8) rectangle ++(0.15,1.0);}
 
\draw [-to](3.65,1) -- (4.05,1) node [pos=0.5, above, font=\footnotesize] {C};

\foreach \X [count=\Z]in {a,b,c}
 {\draw[draw=black, fill=black!20] (3.3+\Z/8, -0.15, 1-\Z/8) rectangle ++(0.15,1.0);}

\draw [-](4.35,1) -- (4.55,1);
\draw [-](4.55,1) -- (4.55,0.5);

\draw [-to](4.55,0.5) -- (4.8,0.5) node [pos=0.5, above, font=\footnotesize] {D};

\foreach \X [count=\Z]in {3,2,1}
 {\node[opacity=1.] at (4.5+\Z/4,0,1-\Z/4) {\includegraphics[width=2.5cm]{images/out\X_wt.png}};}

\draw [-to](5.9,0.5) -- (6.2,0.5) node [pos=0.5, above, font=\footnotesize] {E};

\foreach \X [count=\Z]in {3,2,1}
 {\node[opacity=1.] at (5.9+\Z/4,0,1-\Z/4) {\includegraphics[width=2.5cm]{images/out\X.png}};}
\end{tikzpicture}
}

\caption{The workflow of WCC on a 3-channel input from the Cityscapes dataset. \textbf{A}: Haar-wavelet transform; \textbf{B}: Joint shrinkage, transforming the input into equally sized 1D vectors and a single bit-map to represent the spatial location of the top $k$ entries; \textbf{C}: $1\times1$ convolution over the 1D vectors; \textbf{D}: Inverse shrinkage (zero filling); \textbf{E}: Inverse Haar transform.}
\label{fig:workflow}

\end{figure*}

In this work, we aim to reduce the memory bandwidth and computational cost associated with convolutions performed on intermediate activation maps. To this end, we apply the Haar-wavelet transform to compress the activation maps, in addition to light quantization of 8 bits. Our method is most efficient for scenarios with high-resolution feature maps (\textit{i.e.},~large images, whether in 2D or 3D), where the wavelet compression is most effective. Such cases are mostly encountered in image-to-image tasks like semantic segmentation, depth prediction, image denoising, in-painting, super-resolution, and more. Typically, in such scenarios, the size and memory bandwidth used for the weights are relatively small compared to those used for the features.

\textbf{Convolution in the wavelet domain.}
We focus on the compression of fully-coupled $1\times1$ convolutions, as these are the workhorse of lightweight and efficient architectures like MobileNet \cite{sandler2018mobilenetv2}, ShuffleNet \cite{ma2018shufflenet}, EfficientNet \cite{tan2019efficientnet}, ResNeXt \cite{xie2017resnext}, and ConvNext \cite{liu2022convnext}. All these modern architectures rely on $1\times1$ point-wise convolutions in addition to grouped or depthwise spatial convolutions (\textit{i.e.},~with $3\times3$ or larger kernels), which comprise a small part of the computational effort in the network---the $1\times1$ operations dominate the inference cost (see \autoref{apnd:1x1} and \autoref{apnd:mobilenetmac}). Since we focus on computational efficiency, we use the Haar transform, as it is the simplest and most computationally efficient wavelet variant. Indeed, the Haar transform involves binary pooling-like operations, which include only additions, subtractions, and a bit-wise shift. Other types of wavelets are also suitable in our framework while considering the corresponding computational costs---we demonstrate this point in \autoref{apnd:wavelets}.  

The main idea of our method is to transform and compress the input using the wavelet transform prior to the $1\times1$ convolution, then apply it in the wavelet domain on a fraction of the input size. Since the wavelet compression is applied separately on each channel, it commutes with the $1\times1$ convolution. Hence, applying the convolution in the wavelet domain is equivalent to applying it in the spatial domain.    

More precisely, the advantage of the wavelet transforms is their ability to compress images. Denote the Haar transform operator as $\bfH$, \textit{i.e.},~$\bfH\bfx = \mbox{HWT}(\bfx)$. Then, for most natural images $\bfx$ we have
\begin{equation}\label{eq:HaarApprox}
\bfx \approx \bfH^\top \bfT^\top \bfT\bfH\bfx,
\end{equation}
where $\bfT$ is the shrinkage operator in Eq. \eqref{eq:T} (top $k$ extractor). Because of its orthogonality, $\bfH^\top$ is the inverse transform $\mbox{iHWT}$. Our WCC layer is defined by:
\begin{equation}\label{eq:WCC}
\mbox{WCC}(\bfK_{1\times1},\bfx) = \bfH^\top\bfT^\top \bfK_{1\times 1}\bfT\bfH\bfx,
\end{equation}
where here $\bfx$ is a feature tensor and $\bfK_{1\times 1}$ is a learned $1\times 1$ convolution operator. The workflow is illustrated in \autoref{fig:workflow}, and an explicit algorithm appears in \autoref{apnd:pseudocode}. The convolution operates on the compressed domain, hence, if $\bfT$ can significantly reduce the dimensions of the channels without loosing significant information, this leads to major savings in computations. Note that $\bfH$ and $\bfT$ operate on all channels in the same way, \textit{i.e.},~$\bfT$ extracts the same entries from all channels. We motivate on this next.  

\textbf{Joint hard shrinkage.}
Prior to the $1\times 1$ convolution, the spatial wavelet transform is applied, and we get sparse feature maps. Since the different channels result from the same input image propagated through the network, they typically include patterns and edges at similar locations. Hence the sparsity pattern of their wavelet-domain representation is relatively similar. This idea of redundancy in the channel space is exploited in the works of \cite{han2020ghostnet, eliasof2020multigrid, bae2021dcac}, where parts of the channels are used to represent the others. Therefore, in $\bfT$, we perform a joint shrinkage operation between all the channels, in which we zero and remove the entries with the smallest feature norms across channels, resulting in a compressed representation of the activation maps\footnote{Regardless of the joint sparsity of the channels, some approaches suggest taking the left-upmost part of the wavelet transform for any image, so in principle, the joint sparsity may suffice for a general set of images as well.}. The locations of the non-zeros in the original image are kept in a \emph{single} index list or a bit-map for all the channels in the layer, as they are needed for the inverse transform back to the spatial domain. We also apply light 8-bit quantization to the transformed images to improve the compression rate further. The weights and wavelet-domain activations are quantized using the symmetric scheme described in \autoref{sec: quant-aware-train}.

\textbf{Equivalence of the convolutions.} Our method aims at compression only. Hence, we show that a $1\times 1$ convolution kernel can be applied both in the spatial and in the compressed wavelet domain. By its definition, we can write a $1\times 1$ convolution as a summation over channels. That is:
\begin{equation}
 \bfy = \bfK_{1\times 1}\bfx \Rightarrow  \bfy_i = \textstyle{\sum_{j}k_{ij}\bfx_j},
\end{equation}
where $k_{ij}\in\mathbb{R}$ are the weights of the convolution tensor. Now, suppose we wish to compress the result $\bfy$, through \eqref{eq:HaarApprox}. Because $\bfT$ and $\bfH$ are separable and spatial, we get:
\begin{eqnarray}
\bfy_i \approx \bfH^\top \bfT^\top \bfT\bfH(\bfK_{1\times 1}\bfx)_i  = \bfH^\top \bfT^\top\textstyle{\sum_{j}k_{ij}  \bfT\bfH\bfx_j},  
\end{eqnarray}
where the exact equality holds assuming that $\bfT$ extracts the same entries in both cases. Hence,
\begin{equation}
\bfy \approx \bfH^\top \bfT^\top\bfK_{1\times 1} \bfT\bfH\bfx.
\end{equation}
That is, the wavelet compression operator, which is known to be highly efficient for natural images, commutes with the $1\times 1$ convolution, so the latter can be applied on the compressed signal and get the same result as compressing the result directly. This introduces an opportunity to save computations on the one hand and use more accurate compression on the other. 

The description above is suited for $1\times1$ convolutions only, while many CNN architectures involve spatial convolutions with larger kernels, strides, etc. The main concept is that fully coupled convolutions that mix all channels are expensive, inefficient, and redundant when used with large spatial kernels \cite{ephrath2020leanconvnets}. The spatial mixing can be obtained using separable convolutions at less cost without losing efficiency (see \autoref{apnd:1x1}). Since we aim at saving computations, separating the kernels in the architecture is recommended even before discussing any form of lossy compression \cite{chen2018deeplab,sandler2018mobilenetv2}. Furthermore, separable convolutions (and the Haar transform) can be applied separately in chunks of channels, and the memory bandwidth can be reduced. The $1\times1$ convolution, on the other hand, involves all channels at once, and it is harder to reduce the bandwidth in this case. With our approach, the input for the convolution is compressed, and the bandwidth is reduced (see \autoref{apnd:inferencetime}). 

\textbf{Limitations.} Our approach is best suited for applications where feature maps are of high resolution. This includes various computer vision tasks, as we demonstrate in \autoref{sec:results}. The majority of other works show only classification results, which is a less sensitive task than others for the input's resolution. For example, for ImageNet, images are typically resized to $224\times224$ and down-sampled aggressively to $7\times7$ using a few layers. For such a low resolution, the wavelet shrinkage is not as effective as it is for high-resolution feature maps. On the other hand, quantization by itself is already very effective for classification, hence tackling other tasks in this context may be highly beneficial.  

\section{Results} \label{sec:results}

\begin{figure}[t]
    \centering
    \begin{subfigure}{0.4\textwidth}
        \centering
        \includegraphics[width=0.75\textwidth]{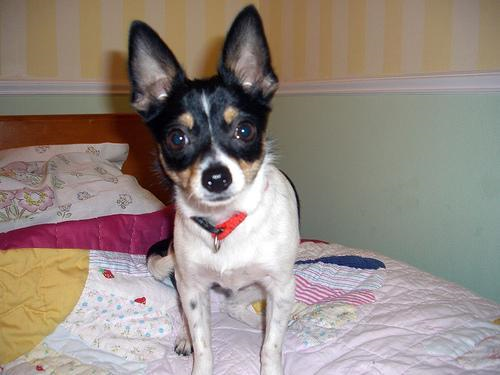}
        \caption{Original input image from ImageNet}
    \end{subfigure}
    \begin{subfigure}{0.4\textwidth}
        \centering
        \includegraphics[width = 0.85\textwidth]{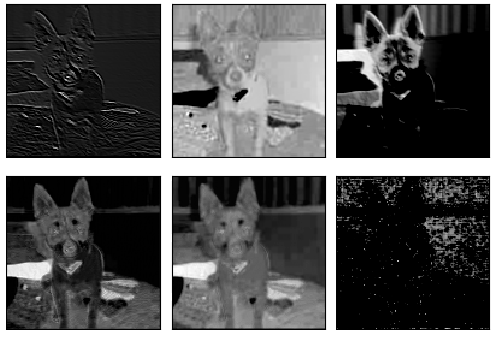}
        \caption{Feature maps from ResNet50 layers.}
    \end{subfigure}
    \begin{subfigure}{0.4\textwidth}
        \centering
        \includegraphics[width=0.85\textwidth]{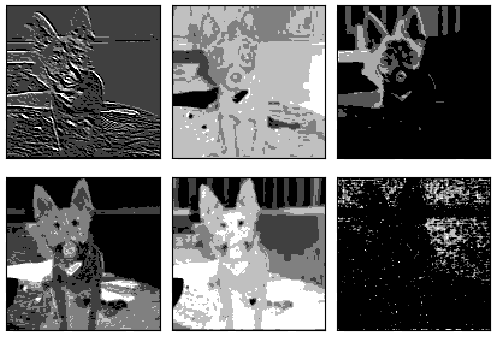}
        \caption{2-bit quantized layers (unsigned).}
    \end{subfigure}
    \begin{subfigure}{0.4\textwidth}
        \centering
        \includegraphics[width=0.85\textwidth]{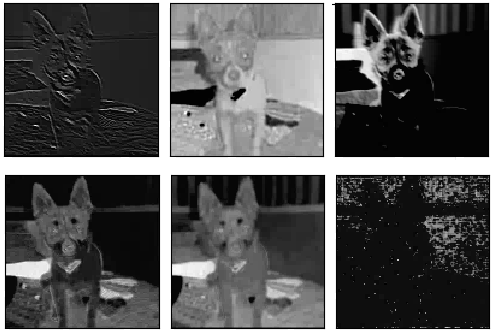}
        \caption{2-bit equivalent wavelet compressed. }
    \end{subfigure}
    \caption{Feature maps from layers 2 and 3 (top and bottom triplets, respectively) of a pre-trained ResNet50 on ImageNet. The maps are compressed with uniform quantization (2-bit) and wavelet compression (25\% shrinkage + 8-bit quantization, equivalent to 2-bit quantization). Clearly, the wavelet compression loses much less information than aggressive quantization.}
    \label{fig:quantization_visualization_compare}
\end{figure}

This section evaluates our proposed WCC layer's performance on four tasks: object detection, semantic segmentation, monocular depth estimation, and super-resolution. The majority of the feature maps in these tasks are relatively large and very well-suited for wavelet compression. 
We use quantization-aware training based on the work of \cite{li2019additive} for the baseline with and without WCC.

As mentioned in \autoref{sec:related}, most quantization works evaluate their performance on image classification. An exception is the work of \cite{nagel2021white}, which shows results for object detection and semantic segmentation, although we were unable to reproduce their results. Therefore, when comparing with their work, we measure the effectiveness of WCC in the relative degradation of the score when compressing the activation further than 8-bit; this is due to the absolute accuracy being strongly dependent on the baseline quantization mechanism and training process.

The general training scheme used in the subsequent experiments is similar to  \cite{li2019additive}. We either start with full-precision pre-trained network weights or train a full-precision network to convergence. Afterward, we gradually reduced the bit rates and the WCC compression factor. We detail the rest of our experimental setup in each section separately. In each network presented in the results, the $1 \times 1$ convolution layers comprise 85\% to 90\% of the total operations (\textit{e.g.},~see \autoref{apnd:mobilenetmac}). We implemented our code using PyTorch \cite{paszke2017automatic}, based on Torchvision and public implementations of the chosen networks. We ran our experiments on NVIDIA 24GB RTX 3090 GPU. The computational cost of each model is measured in Bit-Operations (BOPs, see \autoref{apnd:bops}).

\subsection{Qualitative Compression Assessment}
First, we qualitatively demonstrate the advantage of our approach. We compare standard quantization's mean square error (MSE) to our proposed method on a feature map from a pre-trained MobileNetV3 applied on a batch of $10^3$ images from the ImageNet dataset \cite{krizhevsky2017imagenet}. We consider an effective bit rate range of $[2,8]$. The bit rate of the shrunk wavelet coefficients is 8; multiplying it with the compression ratio yields the effective bit rate (\textit{e.g.},~8 bit and 25\% shrinkage is equivalent to 2 bit). 
\autoref{fig:mse_compare} shows the MSE comparison per effective bit rate, and \autoref{fig:quantization_visualization_compare} presents a visualization comparison of a typical activation map compressed by a standard quantization and our method. Both show that the information loss is more significant in standard quantization.

\subsection{Object Detection}

We apply our method to EfficientDet \cite{tan2020efficientdet} using the EfficientDet-D1 variant.
We train and evaluate the networks on the MS COCO 2017 \cite{lin2014microsoft} object detection dataset. The dataset contains 118K training images and 5K validation images of complex everyday scenes, including detection annotations for 80 classes of common objects in their natural context. Images are of size $\sim640\times480$ pixels, and as defined for the D1 variant, are resized to $640\times640$ pixels. We use a popular publicly available PyTorch EfficientDet implementation\footnote{\url{https://github.com/rwightman/efficientdet-pytorch}} and adopt the training scheme suggested by its authors.
We use the AdamW optimizer, with a learning rate of $10^{-3}$ when initially applying WCC layers and $10^{-4}$ for finetuning. In addition, we apply a learning rate warm-up in the first epoch of training, followed by a cosine learning rate decay. Each compression step is finetuned for 20 to 40 epochs. 
The results are presented in \autoref{table:coco}. We train a quantized network using 4-bit weights and 8-bit activations, used as a baseline, and then further compress the activations using WCC with a 50\% compression factor, which results in comparable BOPs of a 4bit/4bit quantized network. The result significantly surpasses the accuracy achieved by \cite{nagel2021white}, both in the absolute score and in relative degradation from the baseline. 
We continue to show that even when applying a 25\% compression factor, our method outperforms the 4bit/4bit network of \cite{nagel2021white} while reducing the BOPs significantly.

\begin{table}[t]
\small
\begin{minipage}[t]{0.48\textwidth}
\centering
\setlength\tabcolsep{5pt}
\captionsetup{type=table}
\caption{Validation results on MS COCO using EfficientDet-D1. Degradation from the baseline is in parentheses.}
\begin{tabular}{lc|c|cc}
\toprule
Precision      &  Wavelet      & BOPs (B)  & $\uparrow$ mAP \\ 
(W/A)          &  shrinkage    & &                                       \\ 
\midrule
\midrule
FP32            &  None         & 6,144     &  40.08                         \\ 
\midrule
\midrule
\multicolumn{4}{c}{\cite{nagel2021white}}                                  \\
\midrule
4bit / 8bit     &  None         & 280.4     &  35.34                        \\ 
4bit / 4bit     &  None         & 185.8     &  24.70 \textsubscript{(-10.64)} \\ 
\midrule
\midrule
\multicolumn{4}{c}{Our baseline + WCC}                                          \\
\midrule
4bit / 8bit     &  None         & 280.4     &  31.44                        \\ 
4bit / 8bit     &  50\%         & 198.5     &  31.15 \textsubscript{(-0.29)} \\ 
4bit / 8bit     &  25\%         & 155.4     &  27.49 \textsubscript{(-3.95)} \\ 
\bottomrule
\end{tabular}
\label{table:coco}
\end{minipage}
\hfill
\begin{minipage}[t]{0.48\textwidth}
\setlength\tabcolsep{5pt}
\centering
\captionsetup{type=table}
\caption{Validation results for Pascal VOC using DeepLabV3plus(MobileNetV2). Degradation from the baseline is in parentheses.}
\begin{tabular}{lc|c|c}
\toprule
Precision        &  Wavelet    & BOPs (B)    & $\uparrow$ mIoU\\
(W/A)            &  shrinkage  &        \\ 
\midrule
\midrule
\multicolumn{4}{c}{\cite{nagel2021white}} \\
\midrule
FP32             &  None         & 4,534       &  0.729                              \\ 
\midrule
4bit / 8bit      &  None         &   141       &  0.709                              \\
4bit / 4bit      &  None         &   70        &  0.668  \textsubscript{(-0.041)}    \\
\midrule
\midrule
\multicolumn{4}{c}{Our baseline + WCC} \\
\midrule
FP32             &  None         & 4,534       &  0.715                              \\ 
\midrule
4bit / 8bit      &  None         &   141       &  0.675                              \\
4bit / 8bit      &  50\%         &    76       &  0.661 \textsubscript{(-0.014)}     \\ 
4bit / 8bit      &  25\%         &    42       &  0.583 \textsubscript{(-0.092)}     \\ 
4bit / 8bit      &  12.5\%       &    24       &  0.515 \textsubscript{(-0.160)}     \\ 
\bottomrule
\end{tabular}
\label{Table:voc}
\end{minipage}

\end{table}

\subsection{Semantic Segmentation} \label{Sec:segm}

For this experiment, we use the popular DeeplabV3plus \cite{chen2018deeplab} with a MobileNetV2 backbone \cite{sandler2018mobilenetv2}. We evaluated our proposed method on the Cityscapes and Pascal VOC datasets. The Cityscapes dataset \cite{cordts2016cityscapes} contains images of urban street scenes; the images are of size $1024 \times 2048$ with pixel-level annotation of 19 classes. During training, we used a random crop of size $768 \times 768$ and no crop for the validation set. The Pascal VOC \cite{everingham2015pascal} dataset contains images of size $513 \times 513$ with pixel-level annotation of 20 object classes and a background class. We augmented the dataset similarly to \cite{chen2018deeplab}.
We used two configurations for the weights---4 and 8 bits---and for each, we used various compression rates for the activations, both for the standard quantization and WCC. 

\begin{wrapfigure}{r}{7.7cm}
\small
    \centering
    \begin{tikzpicture}
    \begin{axis}[
    width=7.7cm,height=5cm,
    xmajorgrids=true,
    ymajorgrids=true,
    xlabel={Bit-Operations},
    xmin=0, xmax=2400,
    ylabel={mIoU},
    legend pos=south east,
    legend style={nodes={scale=0.5, transform shape}},
    ]
    
    \addplot [draw=lineGreen, mark=o, very thick]
    table{%
    x  y
    2273 0.701
    1213 0.681
    673 0.62
    403 0.552
    };
    \addlegendentry{WCC 8/8 50\%-12.5\%}
    
    \addplot [draw=lineBlue, mark=triangle,  very thick]
    table{%
    x  y
    1136 0.682
    616 0.667
    346 0.621
    211 0.549
    };
    \addlegendentry{WCC 4/8  50\%-12.5\%}
    
    \addplot [draw=lineRed, mark=o, very thick, densely dashed, mark options=solid]
    table{%
    x  y
    2273 0.701
    1705 0.683
    1136 0.173
    };
    \addlegendentry{Quantized 8/8,6,4}
    
    \addplot [draw=lineMaroon, mark=triangle, very thick, densely dashed, mark options=solid]
    table{%
    x  y
    1136 0.682
    852 0.669
    568 0.19
    };
    \addlegendentry{Quantized 4/8,6,4}
    \end{axis}
    \end{tikzpicture}
    
    \caption{Validation results for Cityscapes using DeepLabV3plus(MobileNetV2) compared to BOPs for different bit-rate quantizations and WCC configurations.}
    \label{fig:cityscapescompare}
\end{wrapfigure}
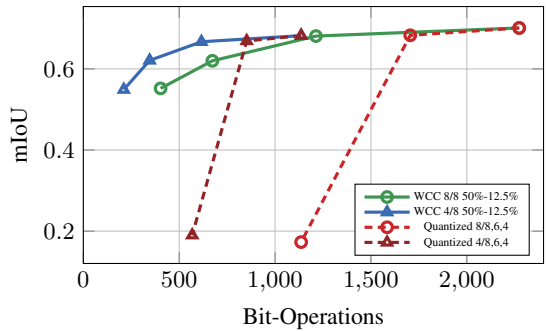

For optimization, we used SGD with momentum $0.9$, weight decay $10^{-4}$, learning rate decay $0.9$, and batch size $16$. For Cityscapes, we trained the full-precision model for 160 epochs with a base learning rate of $10^{-1}$. Each compression step was finetuned for 50 epochs with a learning rate of $10^{-2}$. For Pascal VOC, we trained the full-precision model for 50 epochs with a learning rate of $10^{-2}$. Each compression step was finetuned for 25 epochs with a base learning rate of $2 \cdot 10^{-3}$.

\begin{figure}[t]
    \centering
    \setlength\tabcolsep{1.5pt}
    \captionsetup{subrefformat=parens}
    \begin{tabular}{ccccc}
    
    \subcaptionbox{Input image\label{fig:cs-input}}
        {\includegraphics[width=.24\textwidth]{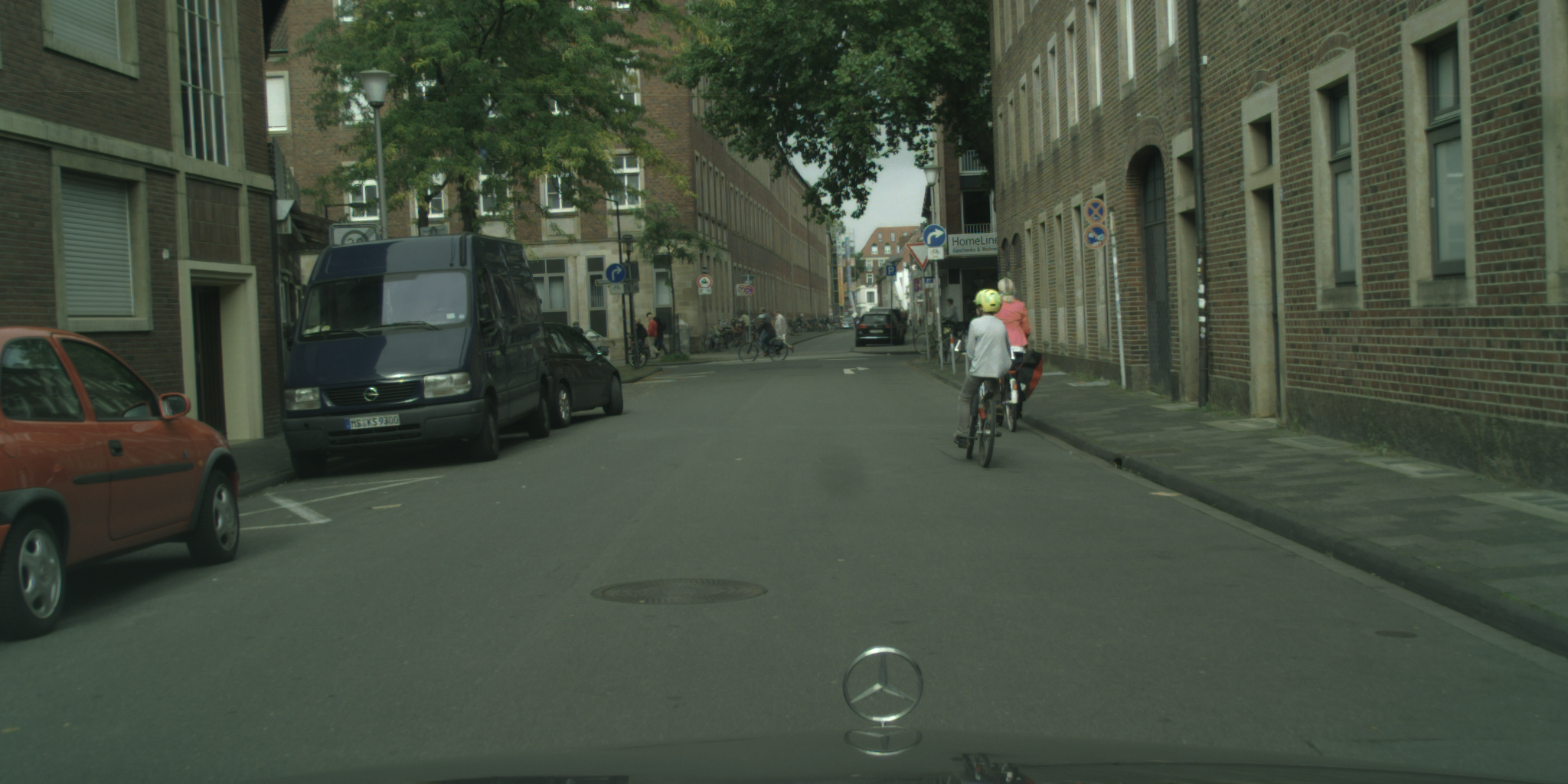}} &
    \subcaptionbox{Ground truth\label{fig:cs-gt}}
        {\includegraphics[width=.24\textwidth]{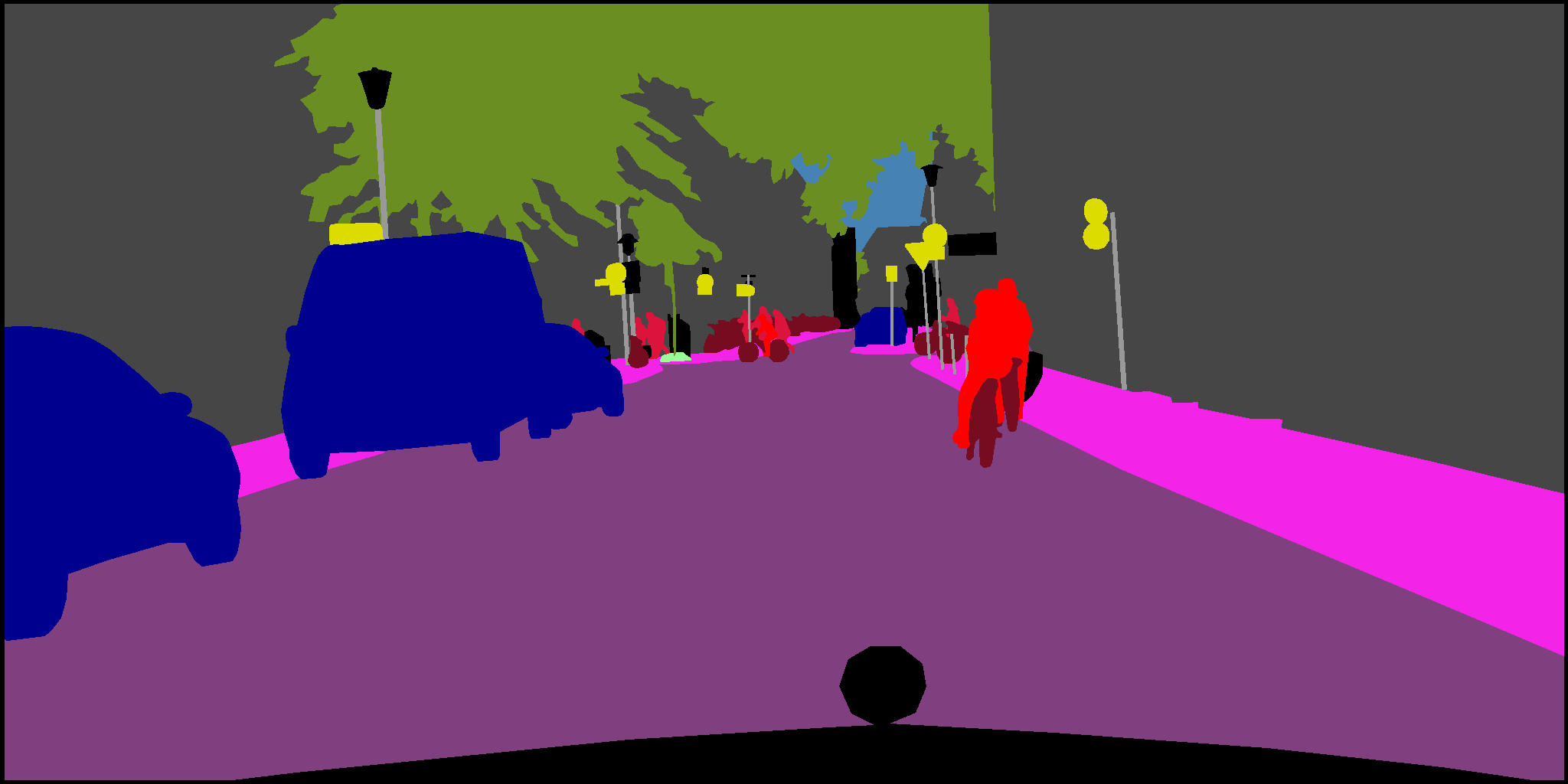}} &
    \subcaptionbox{Baseline 8/8\label{fig:cs-q88}}
        {\includegraphics[width=.24\textwidth]{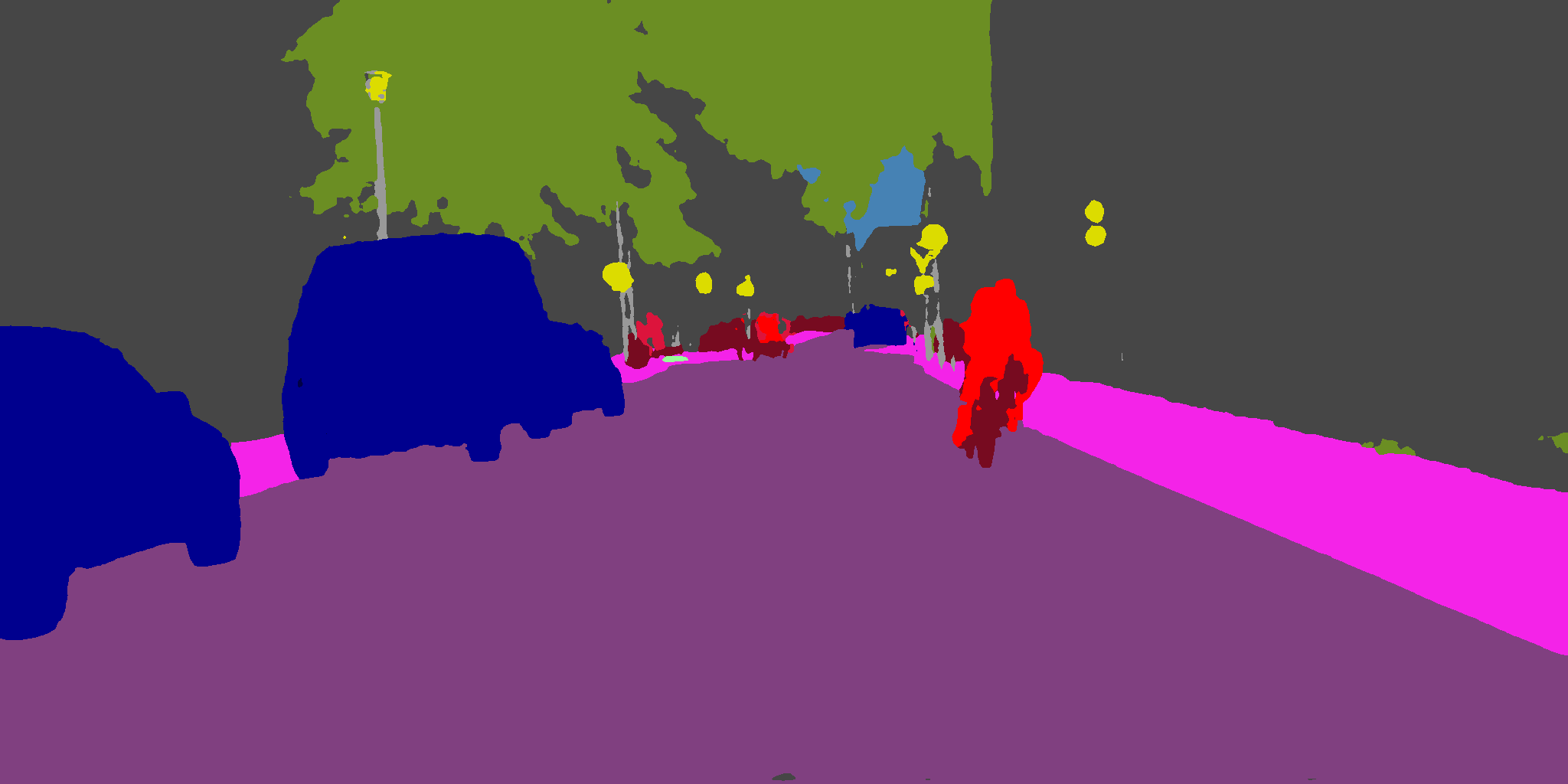}} &
    \subcaptionbox{Baseline 8/4\label{fig:cs-q84}}
        {\includegraphics[width=.24\textwidth]{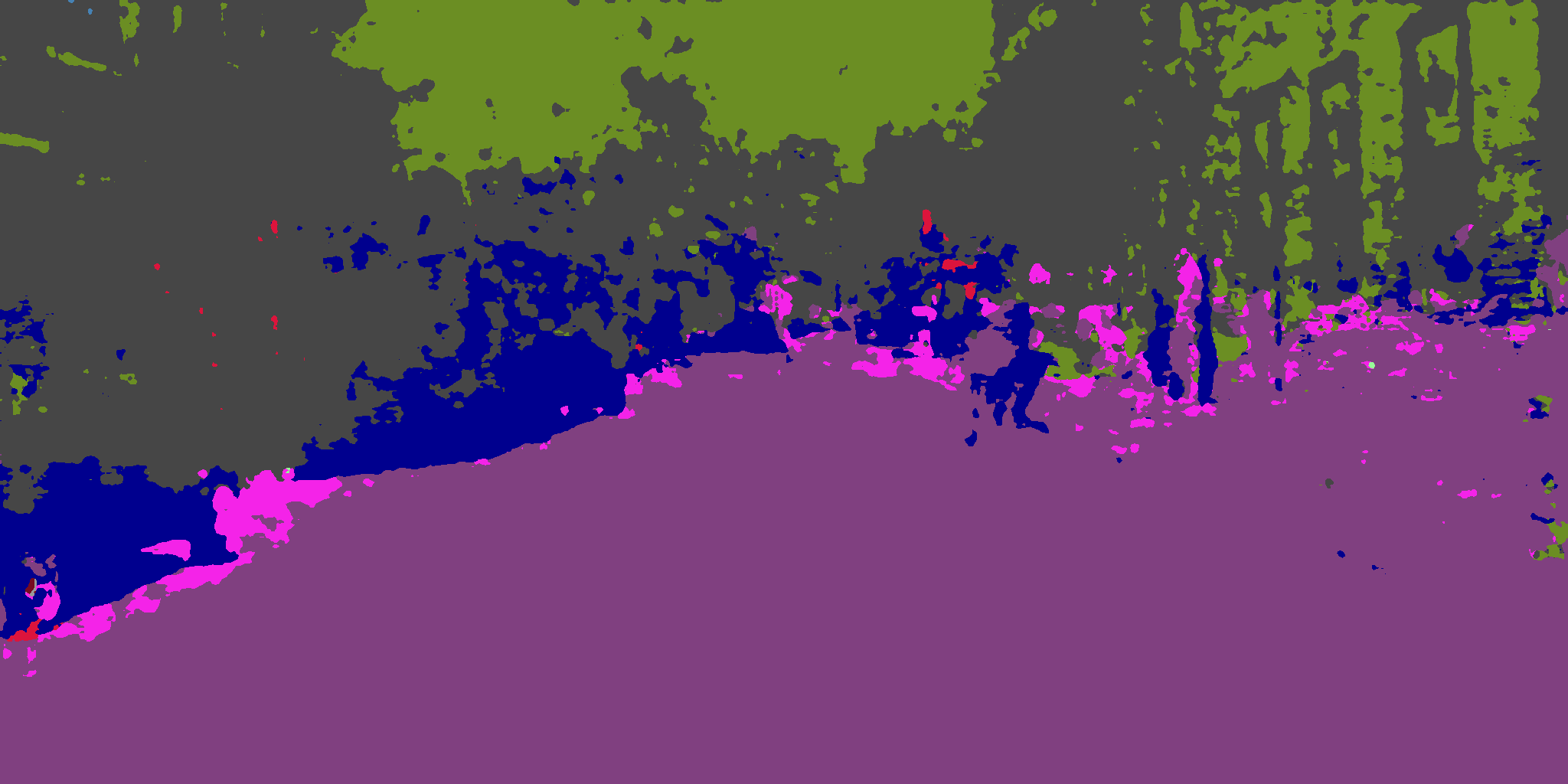}}
    \\
    \\
    \subcaptionbox{WCC 8/8 100\%\label{fig:cs-w100}}
        {\includegraphics[width=.24\textwidth]{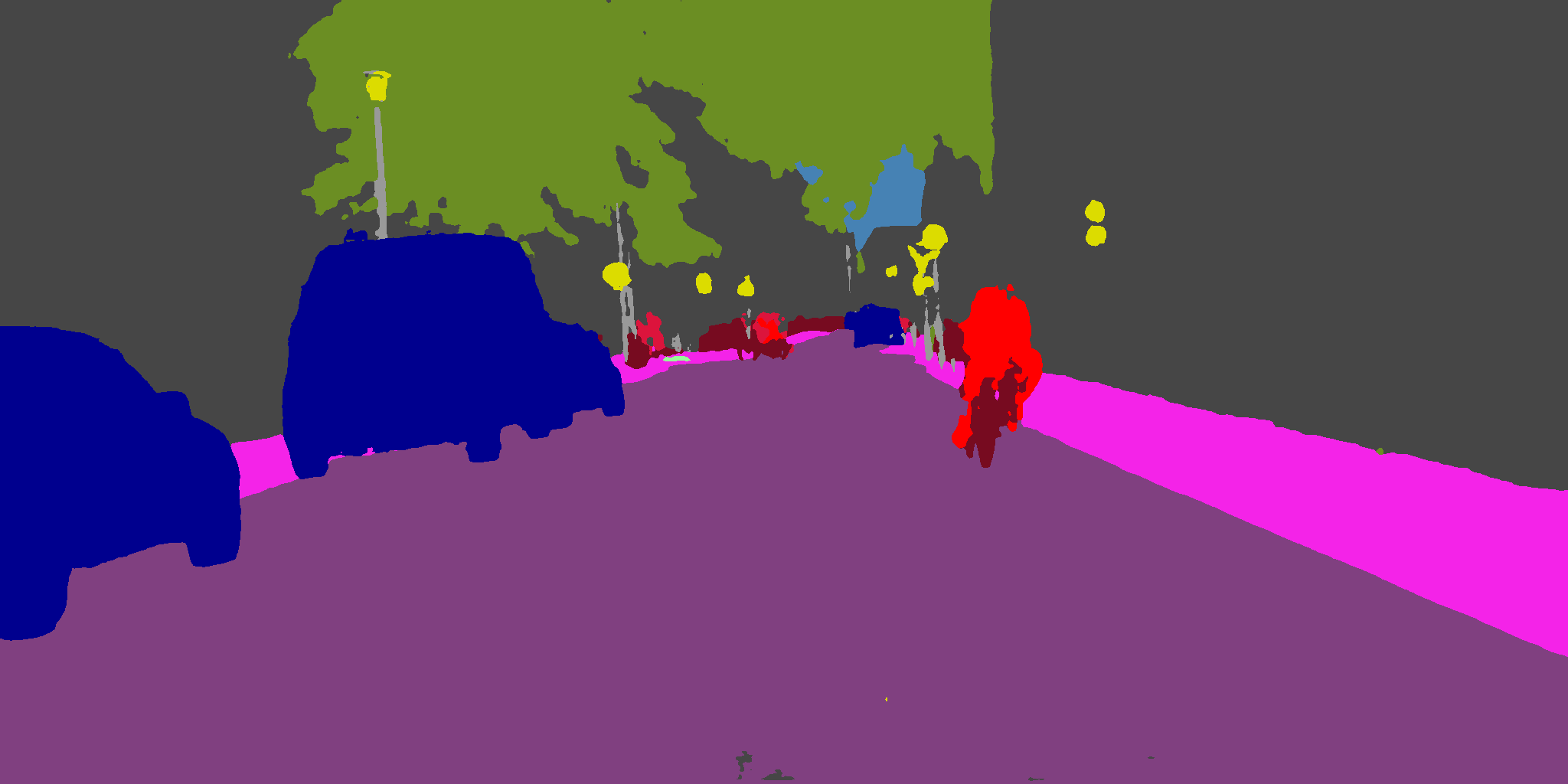}} &
    \subcaptionbox{WCC 8/8 50\%\label{fig:cs-w50}}
        {\includegraphics[width=.24\textwidth]{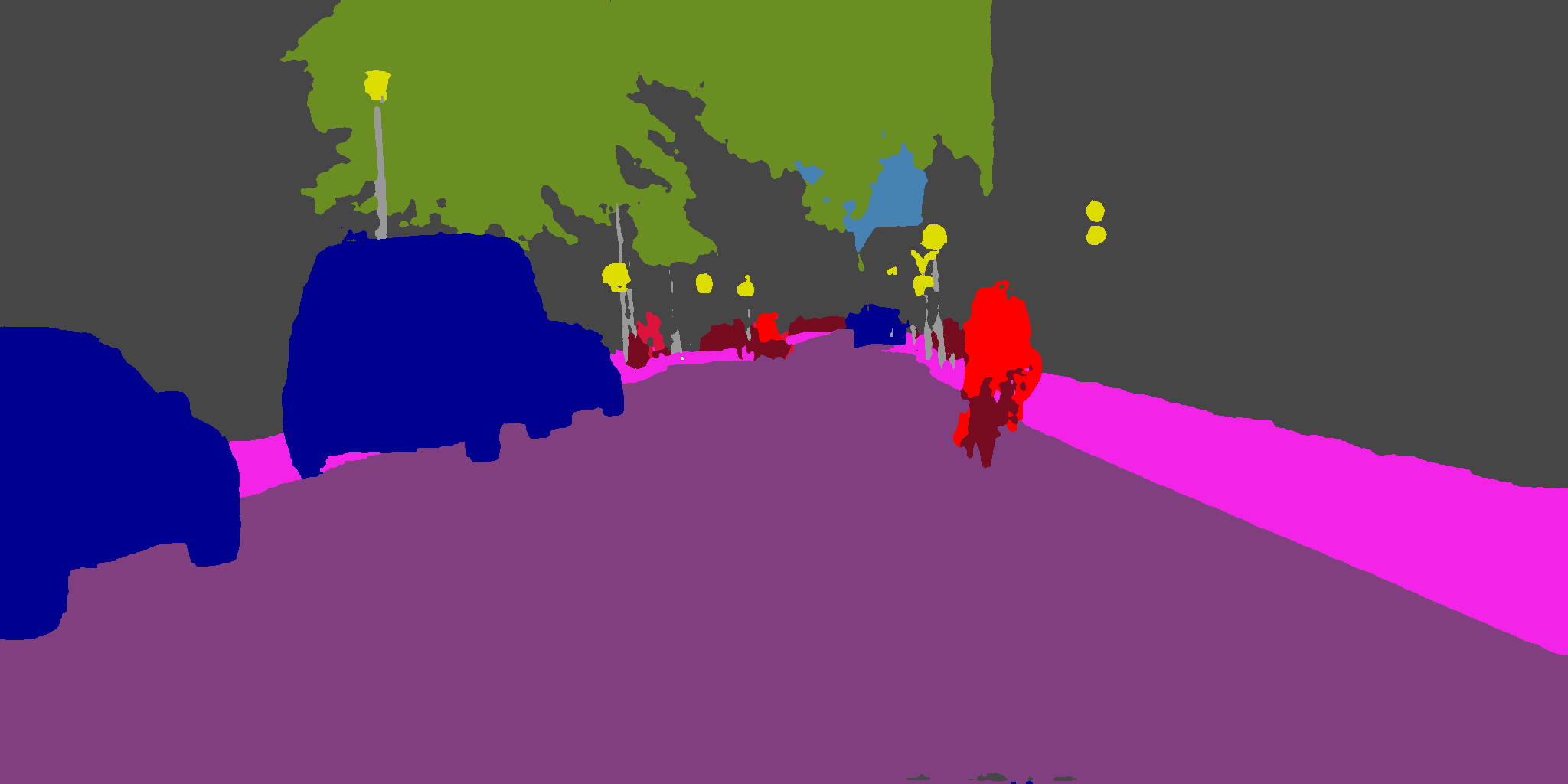}} &
    \subcaptionbox{WCC 8/8 25\%\label{fig:cs-w25}}
        {\includegraphics[width=.24\textwidth]{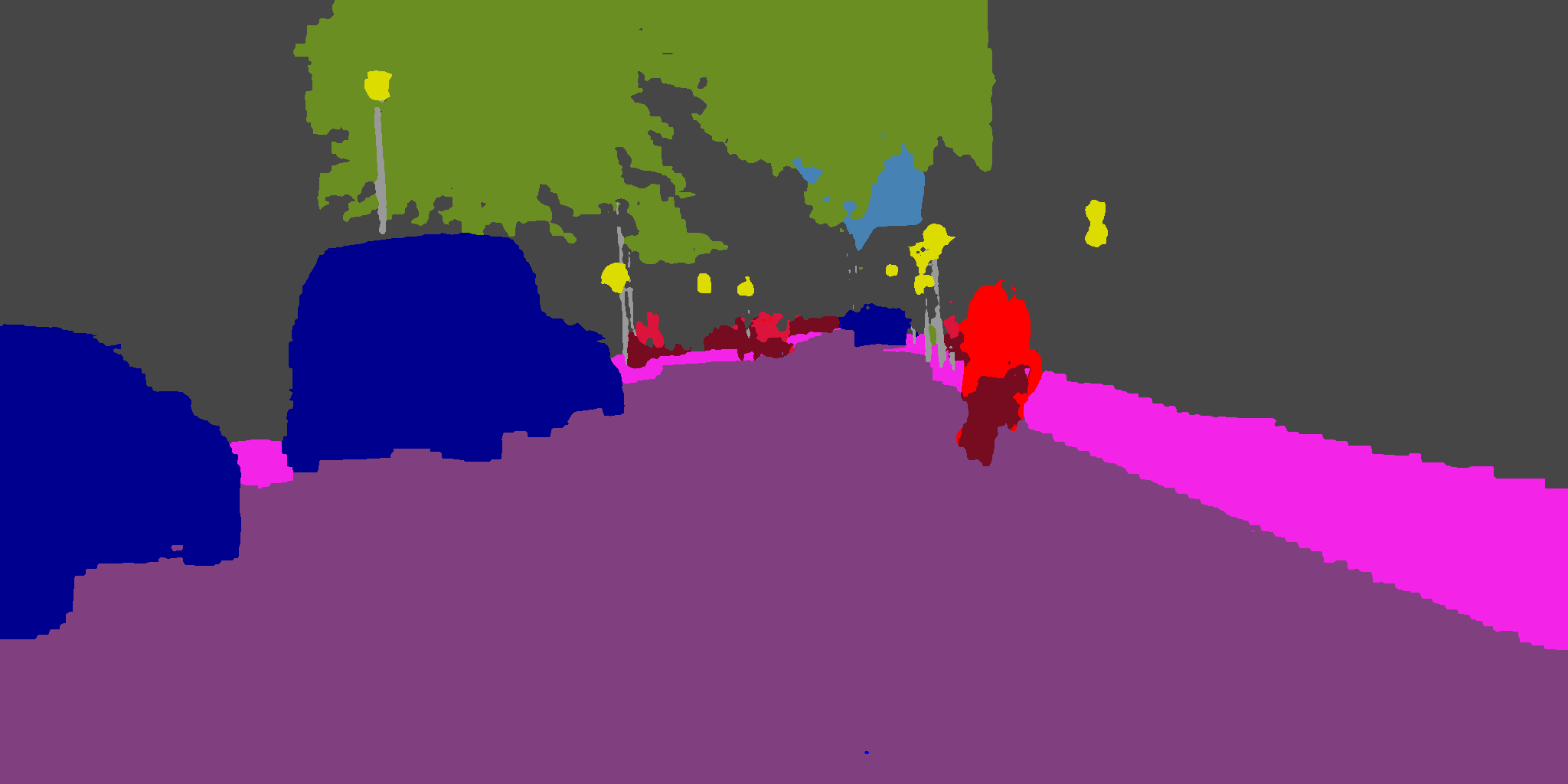}} &
    \subcaptionbox{WCC 8/8 12.5\%\label{fig:cs-w12}}
        {\includegraphics[width=.24\textwidth]{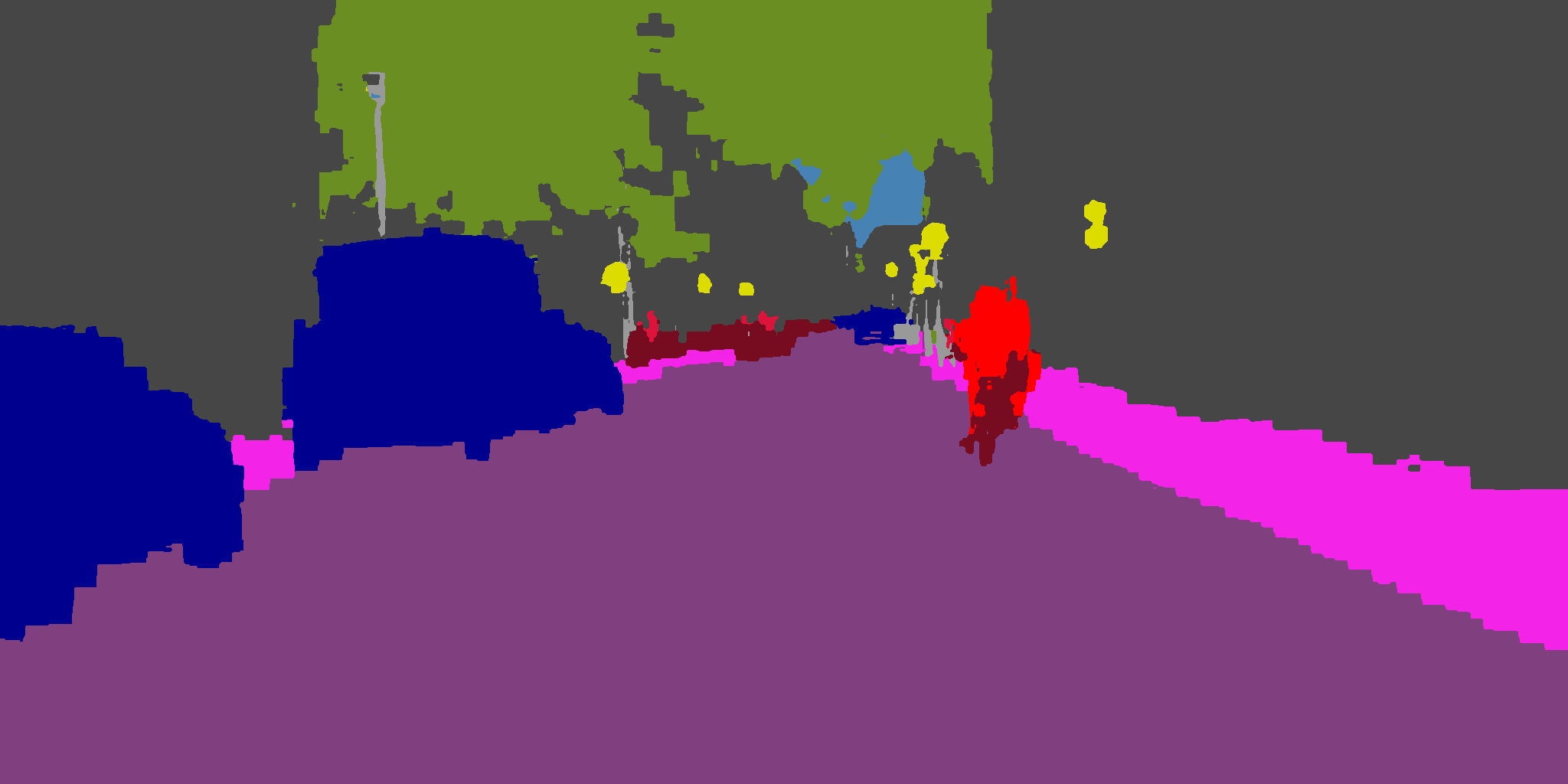}}
    \end{tabular}
    \caption{Cityscapes segmentation results. All networks use weight quantization of 8-bits. \subref{fig:cs-input} input image. \subref{fig:cs-gt} ground truth. \subref{fig:cs-q88}, \subref{fig:cs-q84} normal quantization with 8- and 4-bits activations respectively. \subref{fig:cs-w100}-\subref{fig:cs-w12} WCC with 8-bits activations and shrinkage rate of 100\%, 50\%, 25\% and 12.5\% respectively.}
    \label{Fig:cityscapes}
\end{figure}

\autoref{Table:voc} presents the results for Pascal VOC. In relative degradation, \cite{nagel2021white} experienced a drop of $\sim$0.041 between 4bit/8bit and 4bit/4bit. In contrast, our method experiences a $\sim$0.014 drop when applying 50\% compression, achieving an equivalent score while using a worse baseline. \autoref{fig:cityscapescompare} and \autoref{Fig:cityscapes} compare the results for Cityscapes with different compression configurations. Note that we show compression rates that are computationally equal to 2- and 1-bit, which degrade gracefully.
See \autoref{apnd:segm} for the detailed results, and \autoref{apnd:wavelets} for additional results with different types of wavelets.

\subsection{Monocular Depth Estimation}\label{Sec:monodepth}
We apply our method to Monodepth2 \cite{godard2019digging}, a network architecture for monocular depth estimation---a task of estimating scene depth using a single image.
We evaluated the results on the KITTI dataset \cite{geiger2013IJRR}, containing images of autonomous driving scenarios, each of size $\sim$1241$\times$376 pixels. The train/validation split is the default selected by Monodepth2 (based on \cite{zhou2017unsupervised}), and we evaluate it on the ground truths provided by the KITTI depth benchmark.

\begin{wraptable}{r}{7.3cm}
\small
\centering
\caption{Monodepth2(MobileNetV2) results measured by AbsRel and RMSE.}
\setlength\tabcolsep{3pt}
\begin{tabular}{lc|c|ccccccc}
\toprule
Precision        &  Wavelet      &  BOPs (B)  & $\downarrow$ AbsRel & $\downarrow$ RMSE\\ 
(W/A)          &  shrinkage  & & &      \\ 
\midrule
\midrule
FP32   &  None         & 1,163.6     &  0.093      &  4.022              \\ 
\midrule  
\midrule
8bit / 8bit   &  None         & 133.6     &  0.092       &  4.018       \\ 
8bit / 4bit   &  None         & 99.26     &  0.097       &  4.166       \\ 
8bit / 2bit   & None          & 82.1      & 0.268        & 8.223        \\
\midrule
8bit / 8bit   &  50\%         & 103.9     &  0.098       &  4.217       \\ 
8bit / 8bit   &  25\%         & 88.5      &  0.112       &  4.663       \\ 
8bit / 8bit  &  12.5\%        & 80.8     &  0.131       &  5.046       \\ 
\bottomrule

\end{tabular}
\label{table:monodepth}
\end{wraptable}

We extended the code base to support MobileNetV2 as a backbone, and adapted the depth decoder to use depthwise-separable convolutions for an extra compression factor, without harming the accuracy. All the layers, except the first and final convolutions are compressed. We use the AdamW optimizer, a learning rate of $1.4 \cdot 10^{-4}$, a weight decay of $10^{-3}$ and run each experiment for 20 epochs.

\autoref{table:monodepth} presents a comparison between standard quantization and WCC for Monodepth2. Using WCC, we compress the network to very low bit-rates while having a relatively minimal and more graceful performance degradation. Using WCC layers with 50\% compression resulted in comparable scores to the alternative 8bit/4bit quantization (with similar BOPs).  When applying a compression factor of 25\%, and even 12.5\%, we achieve superior results to the quantized alternative. A qualitative comparison is presented in \autoref{apnd:monodepth}.

\subsection{Super-resolution} \label{Sec:superres}
Another experiment we perform is for the task of super-resolution, which aims to reconstruct a high-resolution image from a single low-resolution image. For this task, we chose the popular EDSR network \cite{lim2017enhanced}, trained on the DIV2K dataset \cite{Agustsson2017CVPRDIV2K}, and used its basic configuration for 2x super-resolution.
DIV2K contains images in 2K resolution, and the model is trained on $48\times48$ random crops of the down-scaled train images as input (which are not down-scaled further throughout the network).
We used the official implementation of EDSR and did not change the training process.

The model consists of three parts: head, body, and tail. In our experience, quantizing the head and tail showed a severe degradation with even the lightest quantization. Therefore, for this experiment, we only compress the body. It is important to mention that this method uses a skip-connection between the head and the tail, so it is to be expected that even aggressive quantization can achieve reasonable results, differently than the other models we show in the paper. 

\begin{wraptable}{r}{7.3cm}
\centering
\small
\setlength\tabcolsep{5pt}
\caption{EDSR compression results for the task of 2x super-resolution. Note that the first row is the reported result from \cite{lim2017enhanced}, while the second is our reproduction of it using separable convolutions.}
\begin{tabular}{lc|c|c}
\toprule
Precision        &  Wavelet      & BOPs (B)   & $\uparrow$ PSNR\\ 
(W/A)            &  shrinkage    & & \\ 
\midrule
\midrule
FP32/FP32 \cite{lim2017enhanced} &  None & 975,838 & 35.03\\
\midrule
\midrule
FP32/FP32        &  None         & 123,674           &  35.02\\ 
\midrule
FP32/FP32        &  50\%         & 70,017   &  34.98\\ 
FP32/FP32        &  25\%         & 42,910   &  34.93\\ 
FP32/FP32        &  12.5\%       & 29,357   &  34.76\\ 
\midrule  % 8 bit weights
\midrule
8bit / 16bit     &  None         & 15,459        &  34.55\\ 
\midrule
8bit / 8bit      &  None         & 7,730         &  34.53\\ 
8bit / 4bit      &  None         & 3,865         &  34.49\\ 
\midrule
8bit / 16bit     &  50\%         & 8,961     &  34.55\\ 
\midrule
8bit / 8bit      &  50\%         & 4,480     &  34.53\\ 
8bit / 8bit      &  25\%         & 2,786     &  34.50\\ 
\bottomrule
\end{tabular}

\label{Table:edsr}
\end{wraptable}

In addition, EDSR uses full $3\times3$ convolutions. Hence, we use the technique described in \autoref{apnd:1x1}, replacing each full-convolution with a set of two convolutions, a depthwise-$3\times3$ followed by a pointwise-$1\times1$. Empirically, we see no change in results in our experiment following this change, and we could still to replicate the results reported by \cite{lim2017enhanced}.

\autoref{Table:edsr} shows the results for two configurations, WCC without quantization and WCC with light quantization (8/16-bit). The BOPs are calculated for the body's convolutions. The first two rows show our reproduction of \cite{lim2017enhanced} results using separable convolutions. The next three rows show the results using our WCC with no quantization, for which the drop in accuracy is minimal. Furthermore, in the quantized results, we can see that adding $50\%$ compression to the 8bit/16bit and 8bit/8bit setups did not influence the accuracy while reducing the costs. Besides that, we again see that WCC achieves better accuracy than quantization alone with aggressive compression (e.g., 8bit/4bit yields a PSNR of 34.49 while 8bit/8bit+50\% yields 34.53). 

\section{Conclusion}
In this work, we presented a new approach for feature map compression, aiming to reduce the memory bandwidth and computational cost of CNNs working at a high resolution. Our approach is based on the classical Haar-wavelet image compression, which has been used for years in standard devices and simple hardware, and in real-time. We save computational costs by applying $1\times1$ convolutions on the shrunk wavelet domain together with light quantization. We show that this approach surpasses aggressive quantization using equivalent compression rates.

\begin{ack}
The research reported in this paper was supported by the Israel Innovation Authority through the Avatar consortium. The authors also thank the Israeli Council for Higher Education (CHE) via the Data Science Research Center and the Lynn and William Frankel Center for Computer Science at BGU. SF is supported by the Kreitman High-tech scholarship.
\end{ack}

\bibliography{refs}

\begin{thebibliography}{10}

\bibitem{Agustsson2017CVPRDIV2K}
Eirikur Agustsson and Radu Timofte.
\newblock Ntire 2017 challenge on single image super-resolution: Dataset and
  study.
\newblock In {\em The IEEE Conference on Computer Vision and Pattern
  Recognition (CVPR) Workshops}, July 2017.

\bibitem{askarihemmat2019u}
MohammadHossein AskariHemmat, Sina Honari, Lucas Rouhier, Christian S.~Perone,
  Julien Cohen-Adad, Yvon Savaria, and Jean-Pierre David.
\newblock U-net fixed point quantization for medical image segmentation.
\newblock In {\em Medical Imaging and Computer Assisted Intervention (MICCAI),
  Hardware Aware Learning Workshop (HAL-MICCAI)}. Springer, 2019.

\bibitem{bae2021dcac}
Seung-Hwan Bae, Hyuk-Jae Lee, and Hyun Kim.
\newblock Dc-ac: Deep correlation-based adaptive compression of feature map
  planes in convolutional neural networks.
\newblock In {\em 2021 IEEE International Symposium on Circuits and Systems
  (ISCAS)}, pages 1--5. IEEE, 2021.

\bibitem{banner2018scalable}
Ron Banner, Itay Hubara, Elad Hoffer, and Daniel Soudry.
\newblock Scalable methods for 8-bit training of neural networks.
\newblock {\em Advances in neural information processing systems}, 31, 2018.

\bibitem{Banner2018ACIQAC}
Ron Banner, Yury Nahshan, Elad Hoffer, and Daniel Soudry.
\newblock Aciq: Analytical clipping for integer quantization of neural
  networks.
\newblock {\em ArXiv}, abs/1810.05723, 2018.

\bibitem{soudry1}
Ron Banner, Yury Nahshan, and Daniel Soudry.
\newblock Post training 4-bit quantization of convolutional networks for
  rapid-deployment.
\newblock {\em Advances in Neural Information Processing Systems}, 32, 2019.

\bibitem{bengio2013estimating}
Yoshua Bengio, Nicholas L{\'e}onard, and Aaron Courville.
\newblock Estimating or propagating gradients through stochastic neurons for
  conditional computation.
\newblock {\em arXiv preprint arXiv:1308.3432}, 2013.

\bibitem{cai2020zeroq}
Yaohui Cai, Zhewei Yao, Zhen Dong, Amir Gholami, Michael~W Mahoney, and Kurt
  Keutzer.
\newblock Zeroq: A novel zero shot quantization framework.
\newblock In {\em Proceedings of the IEEE/CVF Conference on Computer Vision and
  Pattern Recognition}, pages 13169--13178, 2020.

\bibitem{cai2020rethinking}
Zhaowei Cai and Nuno Vasconcelos.
\newblock Rethinking differentiable search for mixed-precision neural networks.
\newblock In {\em Proceedings of the IEEE/CVF Conference on Computer Vision and
  Pattern Recognition}, pages 2349--2358, 2020.

\bibitem{chen2018deeplab}
Liang-Chieh Chen, Yukun Zhu, George Papandreou, Florian Schroff, and Hartwig
  Adam.
\newblock Encoder-decoder with atrous separable convolution for semantic image
  segmentation.
\newblock In {\em Proceedings of the European conference on computer vision
  (ECCV)}, pages 801--818, 2018.

\bibitem{choi2019accurate}
Jungwook Choi, Swagath Venkataramani, Vijayalakshmi Srinivasan, Kailash
  Gopalakrishnan, Zhuo Wang, and Pierce Chuang.
\newblock Accurate and efficient 2-bit quantized neural networks.
\newblock In {\em MLSys}, 2019.

\bibitem{cordts2016cityscapes}
Marius Cordts, Mohamed Omran, Sebastian Ramos, Timo Rehfeld, Markus Enzweiler,
  Rodrigo Benenson, Uwe Franke, Stefan Roth, and Bernt Schiele.
\newblock The cityscapes dataset for semantic urban scene understanding.
\newblock In {\em Proceedings of the IEEE conference on computer vision and
  pattern recognition}, pages 3213--3223, 2016.

\bibitem{daubechies1992ten}
Ingrid Daubechies.
\newblock {\em Ten lectures on wavelets}.
\newblock SIAM, 1992.

\bibitem{dong2019hawqv2}
Zhen Dong, Zhewei Yao, Yaohui Cai, Daiyaan Arfeen, Amir Gholami, Michael~W
  Mahoney, and Kurt Keutzer.
\newblock Hawq-v2: Hessian aware trace-weighted quantization of neural
  networks.
\newblock {\em The Conference on Neural Information Processing Systems
  (NeurIPS)}, 2020.

\bibitem{duan2017sar}
Yiping Duan, Fang Liu, Licheng Jiao, Peng Zhao, and Lu~Zhang.
\newblock {SAR} image segmentation based on convolutional-wavelet neural
  network and markov random field.
\newblock {\em Pattern Recognition}, 64:255--267, 2017.

\bibitem{eliasof2020multigrid}
Moshe Eliasof, Jonathan Ephrath, Lars Ruthotto, and Eran Treister.
\newblock Multigrid-in-channels neural network architectures.
\newblock {\em arXiv preprint arXiv:2011.09128}, 2020.

\bibitem{ephrath2020leanconvnets}
Jonathan Ephrath, Moshe Eliasof, Lars Ruthotto, Eldad Haber, and Eran Treister.
\newblock Leanconvnets: Low-cost yet effective convolutional neural networks.
\newblock {\em IEEE Journal of Selected Topics in Signal Processing},
  14(4):894--904, 2020.

\bibitem{esser2019lsq}
Steven~K Esser, Jeffrey~L McKinstry, Deepika Bablani, Rathinakumar Appuswamy,
  and Dharmendra~S Modha.
\newblock Learned step size quantization.
\newblock {\em The International Conference on Learning Representations
  (ICLR)}, 2020.

\bibitem{everingham2015pascal}
Mark Everingham, SM~Ali Eslami, Luc Van~Gool, Christopher~KI Williams, John
  Winn, and Andrew Zisserman.
\newblock The pascal visual object classes challenge: A retrospective.
\newblock {\em International journal of computer vision}, 111(1):98--136, 2015.

\bibitem{gal2021swagan}
Rinon Gal, Dana~Cohen Hochberg, Amit Bermano, and Daniel Cohen-Or.
\newblock Swagan: A style-based wavelet-driven generative model.
\newblock {\em ACM Transactions on Graphics (TOG)}, 40(4):1--11, 2021.

\bibitem{geiger2013IJRR}
Andreas Geiger, Philip Lenz, Christoph Stiller, and Raquel Urtasun.
\newblock Vision meets robotics: The kitti dataset.
\newblock {\em International Journal of Robotics Research (IJRR)}, 2013.

\bibitem{godard2019digging}
Cl{\'{e}}ment Godard, Oisin {Mac Aodha}, Michael Firman, and Gabriel~J.
  Brostow.
\newblock Digging into self-supervised monocular depth prediction.
\newblock In {\em The International Conference on Computer Vision (ICCV)},
  October 2019.

\bibitem{guo2016dynamic}
Yiwen Guo, Anbang Yao, and Yurong Chen.
\newblock Dynamic network surgery for efficient dnns.
\newblock {\em Advances in neural information processing systems}, 29, 2016.

\bibitem{han2020ghostnet}
Kai Han, Yunhe Wang, Qi~Tian, Jianyuan Guo, Chunjing Xu, and Chang Xu.
\newblock Ghostnet: More features from cheap operations.
\newblock In {\em Proceedings of the IEEE/CVF Conference on Computer Vision and
  Pattern Recognition}, pages 1580--1589, 2020.

\bibitem{han2015learning}
Song Han, Jeff Pool, John Tran, and William Dally.
\newblock Learning both weights and connections for efficient neural network.
\newblock {\em Advances in neural information processing systems}, 28, 2015.

\bibitem{heinrich2018ternarynet}
Mattias~P Heinrich, Max Blendowski, and Ozan Oktay.
\newblock Ternarynet: faster deep model inference without gpus for medical 3d
  segmentation using sparse and binary convolutions.
\newblock {\em International journal of computer assisted radiology and
  surgery}, 13(9):1311--1320, 2018.

\bibitem{howard2019searching}
Andrew Howard, Mark Sandler, Grace Chu, Liang-Chieh Chen, Bo~Chen, Mingxing
  Tan, Weijun Wang, Yukun Zhu, Ruoming Pang, Vijay Vasudevan, et~al.
\newblock Searching for {MobileNetV3}.
\newblock In {\em Proceedings of the IEEE/CVF International Conference on
  Computer Vision}, pages 1314--1324, 2019.

\bibitem{huang2017wavelet}
Huaibo Huang, Ran He, Zhenan Sun, and Tieniu Tan.
\newblock Wavelet-srnet: A wavelet-based cnn for multi-scale face super
  resolution.
\newblock In {\em Proceedings of the IEEE International Conference on Computer
  Vision}, pages 1689--1697, 2017.

\bibitem{hubara2017quantized}
Itay Hubara, Matthieu Courbariaux, Daniel Soudry, Ran El-Yaniv, and Yoshua
  Bengio.
\newblock Quantized neural networks: Training neural networks with low
  precision weights and activations.
\newblock {\em Journal of Machine Learning Research}, 18:187:1--187:30, 2017.

\bibitem{jung2019learning}
Sangil Jung, Changyong Son, Seohyung Lee, Jinwoo Son, Jae-Joon Han, Youngjun
  Kwak, Sung~Ju Hwang, and Changkyu Choi.
\newblock Learning to quantize deep networks by optimizing quantization
  intervals with task loss.
\newblock In {\em Proceedings of the IEEE/CVF Conference on Computer Vision and
  Pattern Recognition}, pages 4350--4359, 2019.

\bibitem{kim2019qkd}
Jangho Kim, Yash Bhalgat, Jinwon Lee, Chirag Patel, and Nojun Kwak.
\newblock Qkd: Quantization-aware knowledge distillation.
\newblock {\em arXiv preprint arXiv:1911.12491}, 2019.

\bibitem{kingsbury1998dual}
Nick Kingsbury.
\newblock The dual-tree complex wavelet transform: a new efficient tool for
  image restoration and enhancement.
\newblock In {\em 9th European Signal Processing Conference (EUSIPCO 1998)},
  pages 1--4. IEEE, 1998.

\bibitem{kirillov2020pointrend}
Alexander Kirillov, Yuxin Wu, Kaiming He, and Ross Girshick.
\newblock Pointrend: Image segmentation as rendering.
\newblock In {\em Proceedings of the IEEE/CVF conference on computer vision and
  pattern recognition}, pages 9799--9808, 2020.

\bibitem{krizhevsky2017imagenet}
Alex Krizhevsky, Ilya Sutskever, and Geoffrey~E Hinton.
\newblock Imagenet classification with deep convolutional neural networks.
\newblock {\em Communications of the ACM}, 60(6):84--90, 2017.

\bibitem{lecun2015deep}
Yann LeCun, Yoshua Bengio, and Geoffrey Hinton.
\newblock Deep learning.
\newblock {\em nature}, 521(7553):436--444, 2015.

\bibitem{li2017trainingquantnets}
Hao Li, Soham De, Zheng Xu, Christoph Studer, Hanan Samet, and Tom Goldstein.
\newblock Training quantized nets: A deeper understanding.
\newblock In {\em Proceedings of the 31st International Conference on Neural
  Information Processing Systems}, pages 5813--5823, 2017.

\bibitem{li2019additive}
Yuhang Li, Xin Dong, and Wei Wang.
\newblock Additive powers-of-two quantization: An efficient non-uniform
  discretization for neural networks.
\newblock {\em The International Conference on Learning Representations
  (ICLR)}, 2020.

\bibitem{lim2017enhanced}
Bee Lim, Sanghyun Son, Heewon Kim, Seungjun Nah, and Kyoung Mu~Lee.
\newblock Enhanced deep residual networks for single image super-resolution.
\newblock In {\em Proceedings of the IEEE conference on computer vision and
  pattern recognition workshops}, pages 136--144, 2017.

\bibitem{lin2014microsoft}
Tsung-Yi Lin, Michael Maire, Serge Belongie, James Hays, Pietro Perona, Deva
  Ramanan, Piotr Doll{\'a}r, and C~Lawrence Zitnick.
\newblock Microsoft coco: Common objects in context.
\newblock In {\em European conference on computer vision}, pages 740--755.
  Springer, 2014.

\bibitem{liu2018multi}
Pengju Liu, Hongzhi Zhang, Kai Zhang, Liang Lin, and Wangmeng Zuo.
\newblock Multi-level wavelet-cnn for image restoration.
\newblock In {\em Proceedings of the IEEE conference on computer vision and
  pattern recognition workshops}, pages 773--782, 2018.

\bibitem{liu2021zero}
Yuang Liu, Wei Zhang, and Jun Wang.
\newblock Zero-shot adversarial quantization.
\newblock In {\em Proceedings of the IEEE/CVF Conference on Computer Vision and
  Pattern Recognition}, pages 1512--1521, 2021.

\bibitem{liu2022convnext}
Zhuang Liu, Hanzi Mao, Chao-Yuan Wu, Christoph Feichtenhofer, Trevor Darrell,
  and Saining Xie.
\newblock A convnet for the 2020s.
\newblock In {\em Proceedings of the IEEE/CVF Conference on Computer Vision and
  Pattern Recognition}, pages 11976--11986, 2022.

\bibitem{louizos2018relaxed}
Christos Louizos, Matthias Reisser, Tijmen Blankevoort, Efstratios Gavves, and
  Max Welling.
\newblock Relaxed quantization for discretized neural networks.
\newblock {\em arXiv preprint arXiv:1810.01875}, 2018.

\bibitem{ma2018shufflenet}
Ningning Ma, Xiangyu Zhang, Hai-Tao Zheng, and Jian Sun.
\newblock Shufflenet v2: Practical guidelines for efficient cnn architecture
  design.
\newblock In {\em Proceedings of the European conference on computer vision
  (ECCV)}, pages 116--131, 2018.

\bibitem{DFQ}
Markus Nagel, Mart~van Baalen, Tijmen Blankevoort, and Max Welling.
\newblock Data-free quantization through weight equalization and bias
  correction.
\newblock In {\em Proceedings of the IEEE/CVF International Conference on
  Computer Vision}, pages 1325--1334, 2019.

\bibitem{nagel2021white}
Markus Nagel, Marios Fournarakis, Rana~Ali Amjad, Yelysei Bondarenko, Mart van
  Baalen, and Tijmen Blankevoort.
\newblock A white paper on neural network quantization.
\newblock {\em arXiv preprint arXiv:2106.08295}, 2021.

\bibitem{paszke2017automatic}
Adam Paszke, Sam Gross, Soumith Chintala, Gregory Chanan, Edward Yang, Zachary
  DeVito, Zeming Lin, Alban Desmaison, Luca Antiga, and Adam Lerer.
\newblock Automatic differentiation in pytorch.
\newblock In {\em Advances in Neural Information Processing Systems}, 2017.

\bibitem{porwik2004haar}
Piotr Porwik and Agnieszka Lisowska.
\newblock The haar-wavelet transform in digital image processing: its status
  and achievements.
\newblock {\em Machine graphics and vision}, 13(1/2):79--98, 2004.

\bibitem{rabbani2002jpeg2000}
Majid Rabbani.
\newblock Jpeg2000: Image compression fundamentals, standards and practice.
\newblock {\em Journal of Electronic Imaging}, 11(2):286, 2002.

\bibitem{sandler2018mobilenetv2}
Mark Sandler, Andrew Howard, Menglong Zhu, Andrey Zhmoginov, and Liang-Chieh
  Chen.
\newblock Mobilenetv2: Inverted residuals and linear bottlenecks.
\newblock In {\em Proceedings of the IEEE conference on computer vision and
  pattern recognition}, pages 4510--4520, 2018.

\bibitem{sun2021mwq}
Qigong Sun, Yan Ren, Licheng Jiao, Xiufang Li, Fanhua Shang, and Fang Liu.
\newblock {MWQ}: Multiscale wavelet quantized neural networks.
\newblock {\em arXiv preprint arXiv:2103.05363}, 2021.

\bibitem{tan2019efficientnet}
Mingxing Tan and Quoc Le.
\newblock Efficientnet: Rethinking model scaling for convolutional neural
  networks.
\newblock In {\em International Conference on Machine Learning}, pages
  6105--6114. PMLR, 2019.

\bibitem{tan2021efficientnetv2}
Mingxing Tan and Quoc Le.
\newblock {EfficientnetV2}: Smaller models and faster training.
\newblock In {\em International Conference on Machine Learning}, pages
  10096--10106. PMLR, 2021.

\bibitem{tan2020efficientdet}
Mingxing Tan, Ruoming Pang, and Quoc~V Le.
\newblock Efficientdet: Scalable and efficient object detection.
\newblock In {\em Proceedings of the IEEE/CVF conference on computer vision and
  pattern recognition}, pages 10781--10790, 2020.

\bibitem{tang2019towards}
Zhiqiang Tang, Xi~Peng, Kang Li, and Dimitris~N Metaxas.
\newblock Towards efficient u-nets: A coupled and quantized approach.
\newblock {\em IEEE transactions on pattern analysis and machine intelligence},
  42(8):2038--2050, 2019.

\bibitem{tung2018deep}
Frederick Tung and Greg Mori.
\newblock Deep neural network compression by in-parallel pruning-quantization.
\newblock {\em IEEE transactions on pattern analysis and machine intelligence},
  42(3):568--579, 2018.

\bibitem{mpd}
S.~Uhlich, L.~Mauch, F.~Cardinaux, K.~Yoshiyama, J.A. Garcia, S.~Tiedemann,
  T.~Kemp, and A.~Nakamura.
\newblock Mixed precision {DNNs}: All you need is a good parametrization.
\newblock {\em The International Conference on Learning Representations
  (ICLR)}, 2020.

\bibitem{vanhoucke2014learningvisrep}
Vincent Vanhoucke.
\newblock Learning visual representations at scale.
\newblock {\em ICLR invited talk}, 1:2, 2014.

\bibitem{vyas2018multiscale}
Aparna Vyas, Soohwan Yu, and Joonki Paik.
\newblock {\em Multiscale transforms with application to image processing}.
\newblock Springer, 2018.

\bibitem{wang2017factorizedconvs}
Min Wang, Baoyuan Liu, and Hassan Foroosh.
\newblock Factorized convolutional neural networks.
\newblock In {\em Proceedings of the IEEE International Conference on Computer
  Vision Workshops}, pages 545--553, 2017.

\bibitem{wang2020differentiable}
Ying Wang, Yadong Lu, and Tijmen Blankevoort.
\newblock Differentiable joint pruning and quantization for hardware
  efficiency.
\newblock In {\em European Conference on Computer Vision}, pages 259--277.
  Springer, 2020.

\bibitem{williams2018wavelet}
Travis Williams and Robert Li.
\newblock Wavelet pooling for convolutional neural networks.
\newblock In {\em International Conference on Learning Representations}, 2018.

\bibitem{wolter2020neural}
Moritz Wolter, Shaohui Lin, and Angela Yao.
\newblock Neural network compression via learnable wavelet transforms.
\newblock In {\em International Conference on Artificial Neural Networks},
  pages 39--51. Springer, 2020.

\bibitem{xie2017resnext}
Saining Xie, Ross Girshick, Piotr Doll{\'a}r, Zhuowen Tu, and Kaiming He.
\newblock Aggregated residual transformations for deep neural networks.
\newblock In {\em Proceedings of the IEEE conference on computer vision and
  pattern recognition}, pages 1492--1500, 2017.

\bibitem{xu2018quantization}
Xiaowei Xu, Qing Lu, Lin Yang, Sharon Hu, Danny Chen, Yu~Hu, and Yiyu Shi.
\newblock Quantization of fully convolutional networks for accurate biomedical
  image segmentation.
\newblock In {\em Proceedings of the IEEE conference on computer vision and
  pattern recognition}, pages 8300--8308, 2018.

\bibitem{yamamoto2021learnable}
Kohei Yamamoto.
\newblock Learnable companding quantization for accurate low-bit neural
  networks.
\newblock In {\em Proceedings of the IEEE/CVF Conference on Computer Vision and
  Pattern Recognition}, pages 5029--5038, 2021.

\bibitem{yao2021hawq}
Zhewei Yao, Zhen Dong, Zhangcheng Zheng, Amir Gholami, Jiali Yu, Eric Tan,
  Leyuan Wang, Qijing Huang, Yida Wang, Michael Mahoney, et~al.
\newblock Hawq-v3: Dyadic neural network quantization.
\newblock In {\em International Conference on Machine Learning}, pages
  11875--11886. PMLR, 2021.

\bibitem{yin2019understanding}
Penghang Yin, Jiancheng Lyu, Shuai Zhang, Stanley Osher, Yingyong Qi, and Jack
  Xin.
\newblock Understanding straight-through estimator in training activation
  quantized neural nets.
\newblock {\em International Conference on Learning Representations, {(ICLR)}},
  2019.

\bibitem{zhang2018lqnets}
Dongqing Zhang, Jiaolong Yang, Dongqiangzi Ye, and Gang Hua.
\newblock Lq-nets: Learned quantization for highly accurate and compact deep
  neural networks.
\newblock In {\em Proceedings of the European conference on computer vision
  (ECCV)}, pages 365--382, 2018.

\bibitem{zhou2018dorefanet}
Shuchang Zhou, Yuxin Wu, Zekun Ni, Xinyu Zhou, He~Wen, and Yuheng Zou.
\newblock Dorefa-net: Training low bitwidth convolutional neural networks with
  low bitwidth gradients.
\newblock {\em The International Conference on Machine Learning (ICML)}, 2018.

\bibitem{zhou2017unsupervised}
Tinghui Zhou, Matthew Brown, Noah Snavely, and David~G. Lowe.
\newblock Unsupervised learning of depth and ego-motion from video.
\newblock In {\em Proceedings of the IEEE Conference on Computer Vision and
  Pattern Recognition (CVPR)}, July 2017.

\end{thebibliography}
\bibliographystyle{plain}

\newpage
\appendix

\section{A Note on \texorpdfstring{$1\times 1$}{1x1} point-wise convolutions} \label{apnd:1x1}
In the case when a certain CNN use $3\times 3$ convolution only, one can split it to two convolution, a depthwise-$3\times 3$ and a $1\times 1$ \citep{wang2017factorizedconvs, vanhoucke2014learningvisrep}. Assuming no strides, the depthwise conv involves with $C_{in}  \cdot 3 \cdot 3 \cdot N_W \cdot N_H$ MAC operations, while the $1\times 1$ conv includes $C_{in} \cdot C_{out} \cdot N_W \cdot N_H$ MAC operations ($C_{out}/9$ times more expensive than depthwise). Meaning, for a large enough $C_{out}$, the $3 \times 3$ convolution has about 8-9 times more MAC operations than the depthwise-$3 \times 3$ convolution and $1\times 1$ convolution.

Some models, such as the ones referenced in \autoref{sec:WaveletMethod}, are defined based on that concept. For example, MobilenetV2 consists of residual blocks that perform $1\times 1$, depthwise-$3\times 3$, and an additional $1\times 1$, and for an image input of size $1024\times 2048$ (\textit{e.g.},~cityscapes), the $1\times 1$-conv has a MAC count of 18,022M, while the $3\times 3$-conv has a MAC count of 1,056M (see \autoref{apnd:mobilenetmac}).

A recent paper \cite{liu2022convnext}, which shows the impact of Resnet50 modifications, explores the idea of separable convolutions in sections 2.3 and 2.4 and demonstrate the effectiveness of it. For another example, in our Monodepth2 experiment (\autoref{Sec:monodepth}), converting to separable convolutions resulted in a $70\%$ drop in BOPs, while AbsRel stayed at 0.093 and RMSE went up from 3.97 to 4.02.

In cases where one might not want to use separable convolutions. We note that the $3\times 3$ convolution is internally implemented as a matrix-matrix multiplication using different shifts of the image. Hence, the wavelet transform can be adapted to transform the shifted images as well, with a specialized implementation. This implementation might hurt the effectiveness of the joint shrinkage, although for high-resolution images we expect it to behave similarly to the separable convolutions, as large smooth areas are consistent between slightly shifted copies of the same image.

\section{Explicit WCC algorithm} \label{apnd:pseudocode}
In Alg.~\ref{alg:wcc} below we present a pseudo-code for performing the WCC layer. We note that the Haar wavelet transform can be obtained in-place and there is no real need to allocate new memory for the large intermediate feature maps $X^0_{ll}$ and $Y^0_{ll}$ during WCC. Only the $1\times1$ conv operation requires a memory allocation, but it is applied on the shrunken vectors. That is another advantage of WCC as standard conv cannot be applied in-place and needs an allocation of both the large feature maps.    
% WCC layer pseudo-code is presented in .
\begin{algorithm}
   \caption{Wavelet Compressed Convolution}
   \label{alg:wcc}
\begin{algorithmic}
   \STATE {\bfseries Input:} feature map $X\in\mathbb{R}^{n_w\times n_h\times C}$ of spatial size $n_w\times n_h$ and $C$ channels, convolution kernel $K_{1\times 1}$, wavelet-transform level $d$, compression rate $\gamma$
   \STATE $X^0_{ll} = X$
   \FOR{$i=1$ {\bfseries to} $d$}
   \STATE $X^i_{ll}, X^i_{lh}, X^i_{hl}, X^i_{hh} = \mbox{HWT}(X^{i-1}_{ll})$
   \ENDFOR
   \STATE Let $X_{wt}$ be a concatenation of $X^{i}_{lh}, X^{i}_{hl}, X^{i}_{hh}$ for $i=1\ldots d$ and $X^d_{ll}$ as in \eqref{eq:wavelet_transform}
   \STATE Calculate vector norm along the channel dimension of $X_{wt}$.
   \STATE Define $I$ as the set of indices of the top $\lceil\gamma n_w n_h\rceil$ vectors by norm.
   \STATE $Y_{wt} = Conv(K_{1\times 1}, X[I])$
   \STATE Initialize a zeroed $Y^0_{ll}\in\mathbb{R}^{n_w\times n_h\times C}$, and set $Y^0_{ll}[I] = Y_{wt}$. 
   \FOR{$i=d-1$ {\bfseries to} $0$}
   \STATE $Y^i_{ll} = \mbox{iHWT}(Y^{i+1}_{ll}, Y^{i+1}_{lh}, Y^{i+1}_{hl}, Y^{i+1}_{hh})$.
   \ENDFOR
   \STATE Return $Y^0_{ll}$
\end{algorithmic}
\end{algorithm}

\newpage 
\section{MobilenetV2 MACs}\label{apnd:mobilenetmac}
A full breakdown of MobilenetV2 MAC operations for a single Cityscapes' image input is provided in \autoref{Table:mobilenetbreakdown}.

\begin{table}[ht]
\small
\centering
\caption{In depth breakdown of MobilenetV2 (as a backbone for deeplabv3+) for a single $1024 \times 2048$ input. $K$ and $S$ refer to the size of the symmetric kernels and strides respectively. The first convolution of the network is omitted, since it is a common practice to avoid quantizing it.}

\renewcommand{\arraystretch}{.9}

\begin{tabular}{l|cccccc|cc|r}
\toprule
Module id &  $C_{in}$ & $C_{out}$ & $K$ & Groups & $S$ & Dilation & $H$ & $W$ & MAC    \\ 
\midrule
InvRes1 conv1    & 32  & 32  & 3 & 32  & 1 & 1 & 513 & 1025 &   150,552,864 \\
InvRes1 conv2    & 32  & 16  & 1 & 1   & 1 & 1 & 511 & 1023 &   267,649,536 \\
InvRes2 conv1    & 16  & 96  & 1 & 1   & 1 & 1 & 513 & 1025 &   807,667,200 \\
InvRes2 conv2    & 96  & 96  & 3 & 96  & 2 & 1 & 513 & 1025 &   112,914,648 \\
InvRes2 conv3    & 96  & 24  & 1 & 1   & 1 & 1 & 256 & 512  &   301,989,888 \\
InvRes3 conv1    & 24  & 144 & 1 & 1   & 1 & 1 & 258 & 514  &   458,307,072 \\
InvRes3 conv2    & 144 & 144 & 3 & 144 & 1 & 1 & 258 & 514  &   169,869,312 \\
InvRes3 conv3    & 144 & 24  & 1 & 1   & 1 & 1 & 256 & 512  &   452,984,832 \\
InvRes4 conv1    & 24  & 144 & 1 & 1   & 1 & 1 & 258 & 514  &   458,307,072 \\
InvRes4 conv2    & 144 & 144 & 3 & 144 & 2 & 1 & 256 & 514  &    42,467,328 \\
InvRes4 conv3	 & 144 & 32  & 1 & 1   & 1 & 1 & 128 & 256  &   150,994,944 \\
InvRes5 conv1    & 32  & 192 & 1 & 1   & 1 & 1 & 130 & 258  &   206,069,760 \\
InvRes5 conv2    & 192 & 192 & 3 & 192 & 1 & 1 & 130 & 258  &    56,623,104 \\
InvRes5 conv3	 & 192 & 32  & 1 & 1   & 1 & 1 & 128 & 256  &   201,326,592 \\
InvRes6 conv1    & 32  & 192 & 1 & 1   & 1 & 1 & 130 & 258  &   206,069,760 \\
InvRes6 conv2    & 192 & 192 & 3 & 192 & 1 & 1 & 130 & 258  &    56,623,104 \\
InvRes6 conv3	 & 192 & 32  & 1 & 1   & 1 & 1 & 128 & 256  &   201,326,592 \\
InvRes7 conv1    & 32  & 192 & 1 & 1   & 1 & 1 & 130 & 258  &   206,069,760 \\
InvRes7 conv2    & 192 & 192 & 3 & 192 & 2 & 1 & 130 & 258  &    14,155,776 \\
InvRes7 conv3	 & 192 & 64  & 1 & 1   & 1 & 1 & 64  & 128  &   100,663,296 \\
InvRes8 conv1    & 64  & 384 & 1 & 1   & 1 & 1 & 66  & 130  &   210,862,080 \\
InvRes8 conv2    & 384 & 384 & 3 & 384 & 1 & 1 & 66  & 130  &    28,311,552 \\
InvRes8 conv3	 & 384 & 64  & 1 & 1   & 1 & 1 & 64  & 128  &   201,326,592 \\
InvRes9 conv1    & 64  & 384 & 1 & 1   & 1 & 1 & 66  & 130  &   210,862,080 \\
InvRes9 conv2    & 384 & 384 & 3 & 384 & 1 & 1 & 66  & 130  &    28,311,552 \\
InvRes9 conv3	 & 384 & 64  & 1 & 1   & 1 & 1 & 64  & 128  &   201,326,592 \\
InvRes10 conv1   & 64  & 384 & 1 & 1   & 1 & 1 & 66  & 130  &   210,862,080 \\
InvRes10 conv2   & 384 & 384 & 3 & 384 & 1 & 1 & 66  & 130  &    28,311,552 \\
InvRes10 conv3   & 384 & 64  & 1 & 1   & 1 & 1 & 64  & 128  &   201,326,592 \\
InvRes11 conv1   & 64  & 384 & 1 & 1   & 1 & 1 & 66  & 130  &   210,862,080 \\
InvRes11 conv2   & 384 & 384 & 3 & 384 & 1 & 1 & 66  & 130  &    28,311,552 \\
InvRes11 conv3   & 384 & 96  & 1 & 1   & 1 & 1 & 64  & 128  &   301,989,888 \\
InvRes12 conv1   & 96  & 576 & 1 & 1   & 1 & 1 & 66  & 130  &   474,439,680 \\
InvRes12 conv2   & 576 & 576 & 3 & 576 & 1 & 1 & 66  & 130  &    42,467,328 \\
InvRes12 conv3   & 576 & 96  & 1 & 1   & 1 & 1 & 64  & 128  &   452,984,832 \\
InvRes13 conv1   & 96  & 576 & 1 & 1   & 1 & 1 & 66  & 130  &   474,439,680 \\
InvRes13 conv2   & 576 & 576 & 3 & 576 & 1 & 1 & 66  & 130  &    42,467,328 \\
InvRes13 conv3   & 576 & 96  & 1 & 1   & 1 & 1 & 64  & 128  &   452,984,832 \\
InvRes14 conv1   & 96  & 576 & 1 & 1   & 1 & 1 & 66  & 130  &   474,439,680 \\
InvRes14 conv2   & 576 & 576 & 3 & 576 & 1 & 1 & 66  & 130  &    42,467,328 \\
InvRes14 conv3   & 576 & 160 & 1 & 1   & 1 & 1 & 64  & 128  &   754,974,720 \\
InvRes15 conv1   & 160 & 960 & 1 & 1   & 1 & 1 & 68  & 132  & 1,378,713,600 \\
InvRes15 conv2   & 960 & 960 & 3 & 960 & 1 & 2 & 68  & 132  &    70,778,880 \\
InvRes15 conv3   & 960 & 160 & 1 & 1   & 1 & 1 & 64  & 128  & 1,258,291,200 \\
InvRes16 conv1   & 160 & 960 & 1 & 1   & 1 & 1 & 68  & 132  & 1,378,713,600 \\
InvRes16 conv2   & 960 & 960 & 3 & 960 & 1 & 2 & 68  & 132  &    70,778,880 \\
InvRes16 conv3   & 960 & 160 & 1 & 1   & 1 & 1 & 64  & 128  & 1,258,291,200 \\
InvRes17 conv1   & 160 & 960 & 1 & 1   & 1 & 1 & 68  & 132  & 1,378,713,600 \\
InvRes17 conv2   & 960 & 960 & 3 & 960 & 1 & 2 & 68  & 132  &    70,778,880 \\
InvRes17 conv3   & 960 & 320 & 1 & 1   & 1 & 1 & 64  & 128  & 2,516,582,400 \\
\midrule
Total of $1\times 1$ & & & & & & & & & 18,022,413,312 \\
Total of $3\times 3$ & & & & & & & & & 1,056,190,968  \\
\bottomrule
\end{tabular}
\label{Table:mobilenetbreakdown}
\end{table}

\newpage

\section{Computational Costs in Bit Operations (BOPs)}\label{apnd:bops}
To evaluate the computational cost involved in WCC we use the measure of Bit-Operations (BOPs) \citep{wang2020differentiable, louizos2018relaxed}. First, the number of Multiply-And-Accumulate (MAC) operations in a convolutional layer is given by
\begin{align}
    \textrm{MAC}(\textrm{conv}) = C_{\textrm{in}} \cdot C_{\textrm{out}} \cdot N_W \cdot N_H \cdot K_W \cdot K_H \cdot \textstyle{\frac{1}{S_W\cdot S_H}},
\end{align}
where $C_{\textrm{in}}$ and $C_{\textrm{out}}$ are the number of input and output channels, $(N_W,N_H)$ is the size of the input, $(K_W, K_H)$ is the size of the kernel, and $(S_W, S_H)$ is the stride value. The BOPs count is then
\begin{align}\label{eq:BOPs_def}
    \textrm{BOPs}(\textrm{conv}) = \textrm{MAC}(\textrm{conv}) \cdot b_{w}\cdot b_{a},
\end{align}
where $b_w$ and $b_a$ denote the number of bits used for weight and activations.

As described in \autoref{sec: quant-aware-train}, the Haar transform is separable between the input channels, and can be viewed as four $2\times 2$ convolutions with stride $(2,2)$ and binary weights. Hence, the one-level transform requires
$ 4 \cdot C_{\textrm{in}} \cdot W \cdot H \cdot b_{a} $
BOPs. The transform can be used with more levels of compression explained in \autoref{sec: quant-aware-train}, on down-scaled inputs, resulting in a total of
\begin{align}
     \textstyle{\sum_{l=1}^{L} 4 \cdot C_{\textrm{in}} \cdot N_W \cdot N_H \cdot \frac{1}{4^{l-1}} \cdot b_{a}}
\end{align}
BOPs, where $L$ is the level of compression. Similarly, the inverse-transform result in the same calculation, only with $C_{\textrm{out}}$ in place of $C_{\textrm{in}}$.
To demonstrate the relatively small cost of the compression, consider a $1\times 1$ convolution with $C_{\textrm{in}}=160$, $C_{\textrm{out}}=960$, input size of $(34, 34)$, and quantization $b_w=b_a=8$ (which is part of a network used in \autoref{sec:results}). This layer costs $11,364M$ BOPs. Using a 3 levels wavelet transform and its inverse for this layer results in $54M$ BOPs, a negligible cost which allows for better compression, as we demonstrate next.

\section{Full Segmentation Results} \label{apnd:segm}

\autoref{Table:segmentation} shows the performance and BOPs of each model trained by us in the experiments described in \autoref{Sec:segm}.
In our experience, quantizing with 4bit activations resulted in a sharp drop in results. While other more sophisticated methods experience less of a decline, it is still significant. Using said methods for 8bit with our approach will also result in improved scores for WCC.

\begin{table}[ht]
\centering

\caption{Validation results for semantic segmentation task using DeepLabV3plus with MobileNetV2 as the backbone. Segmentation performance is measured by mean intersection over union (mIoU)}

\begin{tabular}{lc|cc|cc}
\toprule
Precision        &  Wavelet      & \multicolumn{2}{c|}{Cityscapes}  & \multicolumn{2}{c}{Pascal VOC}\\ 
(W/A)          &  shrinkage  & BOPs (B)   & mIoU      & BOPs (B)    & mIoU       \\ 
\midrule
\midrule
FP32   &  None         & 36,377     &  0.717    & 4,534       &  0.715     \\ 
\midrule  % 8 bit weights
\midrule
8bit / 8bit      &  None         &  2,273     &  0.701    &   283       &  0.712     \\ 
8bit / 6bit      &  None         &  1,705     &  0.683    &   212       &  0.678     \\ 
8bit / 4bit      &  None         &  1,136     &  0.173    &   141       &  0.095     \\ 
\midrule
8bit / 8bit      &  50\%         &  1,213     &  0.681    &   150       &  0.675     \\ 
8bit / 8bit      &  25\%         &    673     &  0.620    &    82       &  0.611     \\ 
8bit / 8bit      &  12.5\%       &    403     &  0.552    &    48       &  0.519     \\ 
\midrule  % 4 bit weights
\midrule
4bit / 8bit      &  None         &  1,136     &  0.682    &   141       &  0.675     \\ 
4bit / 6bit      &  None         &    852     &  0.669    &   106       &  0.657     \\ 
4bit / 4bit      &  None         &    568     &  0.190    &    70       &  0.099     \\ 
\midrule
4bit / 8bit      &  50\%         &    616     &  0.667    &    76       &  0.661     \\ 
4bit / 8bit      &  25\%         &    346     &  0.621    &    42       &  0.583     \\ 
4bit / 8bit      &  12.5\%       &    211     &  0.549    &    24       &  0.515     \\ 
\bottomrule
\end{tabular}

\label{Table:segmentation}
\end{table}

\section{WCC with Different Wavelets} \label{apnd:wavelets}
When considering different wavelets for compression, the added computational cost should also be weighted. Calculating the MAC operations for the transform as explained in \autoref{apnd:bops}, a $2k\times 2k$ kernel represented in $b_w$ bits results in $k^2b_w$ times the BOPs for the same input compared to the Haar transform.
\autoref{Table:ablation_wavelet} compares different WCC layer configurations using several wavelets for Cityscapes semantic segmentation task, $b_w$ was set to 32-bit floating-point for all options.

\begin{table}[ht]
\setlength\tabcolsep{5pt}
\centering
\caption{Validation results for semantic segmentation task using DeepLabV3plus with MobileNetV2 as the backbone. All experiments are using 8bit/8bit quantization.}
\begin{tabular}{lc|c|c|c}
\toprule
Wavelet Type      & Filter Size & 50\% Shrinkage  & 25\% Shrinkage  & 12.5\% Shrinkage     \\
                  &             & mIoU            & mIoU            & mIoU  \\
\midrule
\midrule
Haar                         & $2\times2$ &  0.681  &  0.620  &  0.552 \\ 
Daubechies 2 (db2)           & $4\times4$ &  0.680  &  0.630  &  0.561 \\ 
Daubechies 3 (db3)           & $6\times6$ &  0.678  &  0.629  &  0.560 \\ 
Coiflets 1 (coif1)           & $6\times6$ &  0.676  &  0.637  &  0.562 \\ Biorthogonal 1.3 (bior1.3)   & $6\times6$ &  0.684  &  0.637  &  0.564 \\ 
Biorthogonal 2.2 (bior2.2)   & $6\times6$ &  0.677  &  0.638  &  0.566 \\ 
Symlets 4 (sym4)             & $8\times8$ &  0.675  &  0.629  &  0.565 \\
\bottomrule
\end{tabular}
\label{Table:ablation_wavelet}
\end{table}

\section{Depth Prediction Qualitative Results} \label{apnd:monodepth}
\begin{figure}[ht]
    \centering
    \setlength\tabcolsep{1.5pt}
    \captionsetup{subrefformat=parens}
    \begin{tabular}{ccc}
    
    \subcaptionbox{Input image\label{fig:kitti-input}}
        {\includegraphics[width=.49\textwidth]{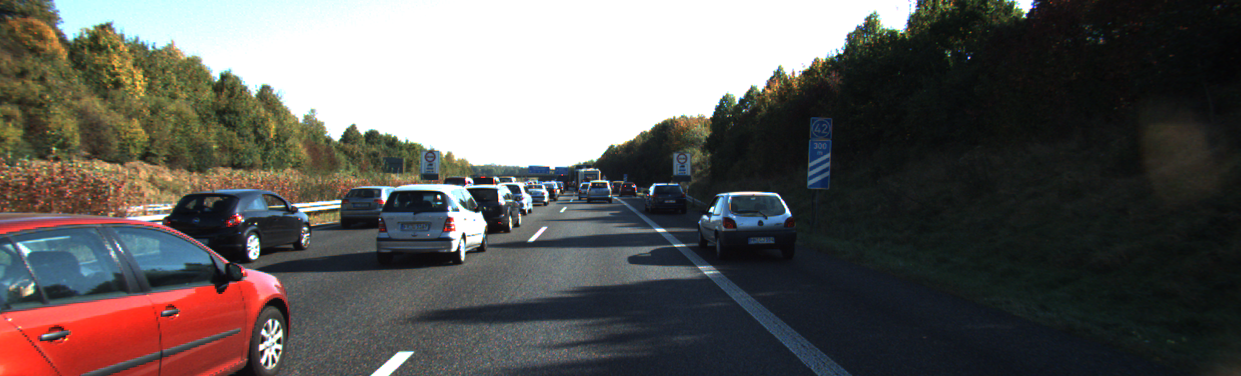}} &
    \subcaptionbox{WCC 8/8 50\%\label{fig:kitti-wt50}}
        {\includegraphics[width=.49\textwidth]{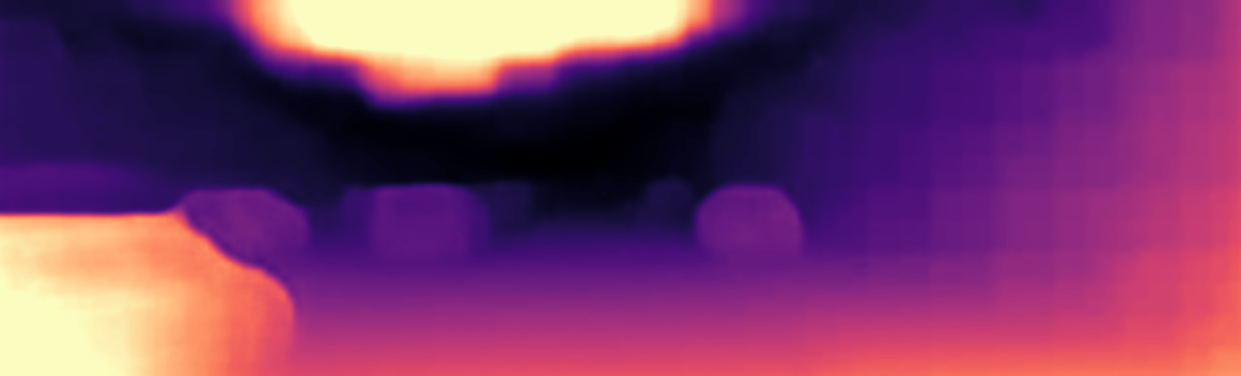}} 
    \\
    \\
    \subcaptionbox{Quantization 8bit / 4bit\label{fig:kitti-q84}}
        {\includegraphics[width=.49\textwidth]{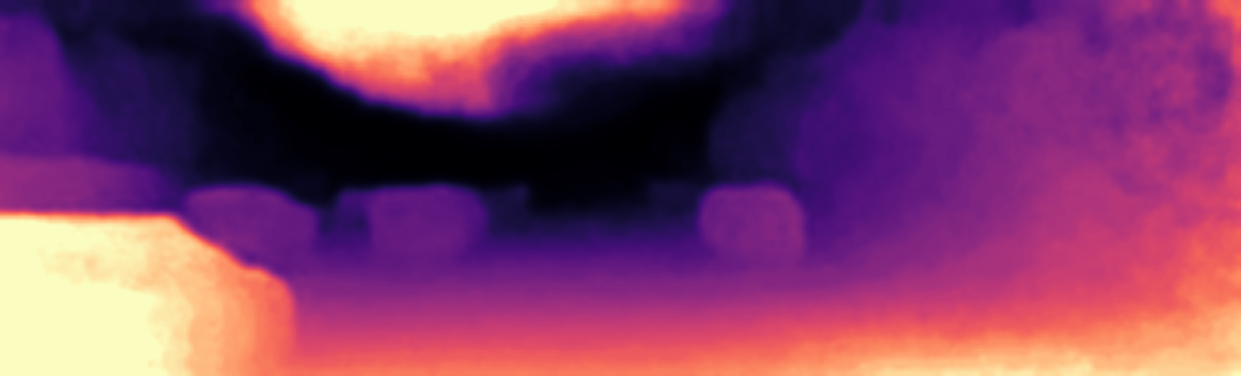}} &
    \subcaptionbox{WCC 8/8 25\%\label{fig:kitti-wt25}}
        {\includegraphics[width=.49\textwidth]{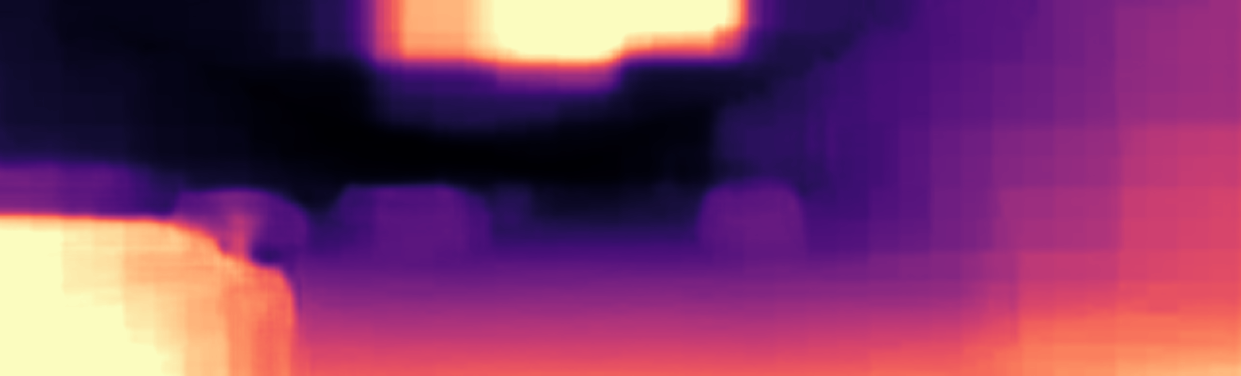}}
    \\
    \\
    \subcaptionbox{Quantization 8bit / 2bit\label{fig:kitti-q82}}
        {\includegraphics[width=.49\textwidth]{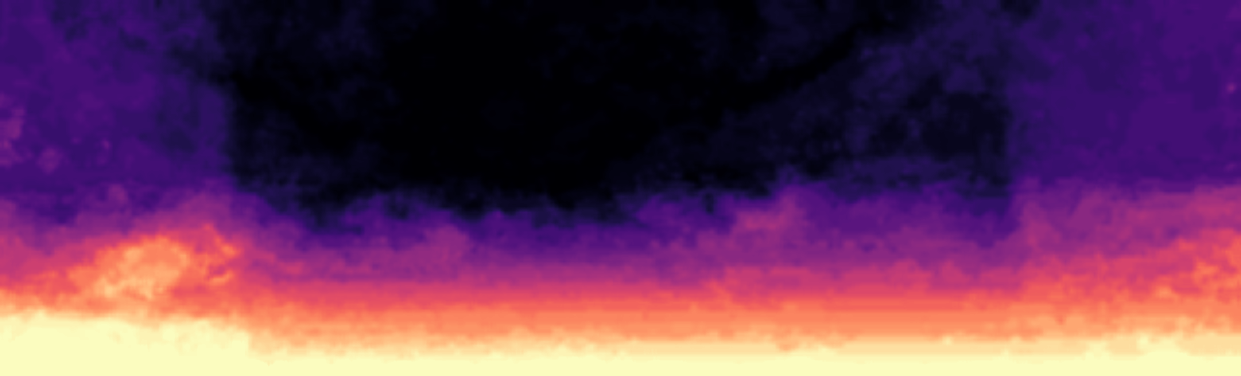}} &
    \subcaptionbox{WCC 8/8 12.5\%\label{fig:kitti-wt125}}
        {\includegraphics[width=.49\textwidth]{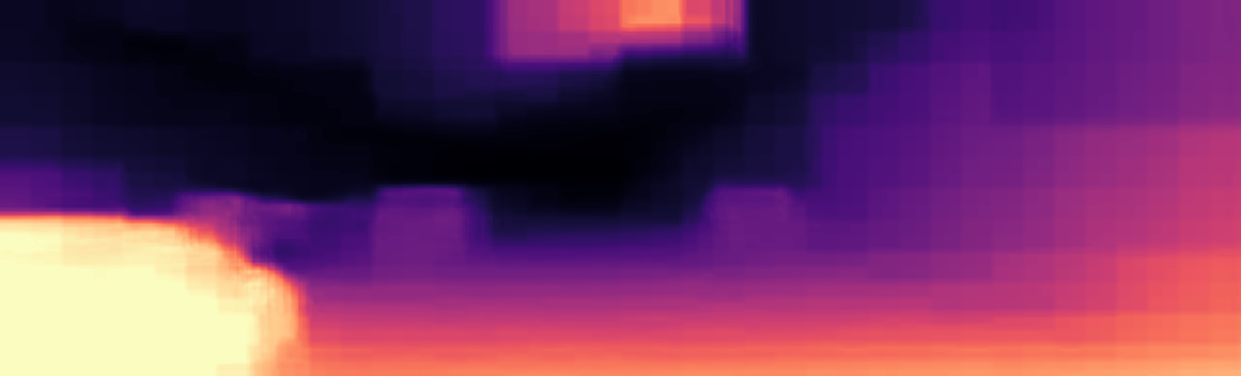}} 
    \end{tabular}
    \caption{Kitti depth estimation prediction examples on Monodepth2. All networks use weight quantization of 8-bits. \subref{fig:kitti-q84}, \subref{fig:kitti-q82} show results for activation quantization of 4bit and 2bit respectively. \subref{fig:kitti-wt50}, \subref{fig:kitti-wt25}, \subref{fig:kitti-wt125} show results for WCC with 50\%, 25\% \& 12.5\% compression factor respectively. }
    \label{fig:kitti}
\end{figure}

Qualitive results for \autoref{Sec:monodepth} are presented in \autoref{fig:kitti}.

\newpage
\section{Inference Times on a GPU using a Custom Implementation}
\label{apnd:inferencetime}
In this paper we followed the common practice of measuring the theoretical speedup of our approach in terms of BOPs, as commonly done in quantization works (e.g., \cite{yao2021hawq}). That is because our aim is to speed up inference times on low-resource devices where BOPs are the main computational bottleneck. However, WCC includes several operations done in tandem which are not straightforward to implement efficiently using existing CNN frameworks. These include the forward and inverse Haar transforms, gather and scatter operations using a single index list, and top-k selection on the pixel-wise norms across channels. To demonstrate that our approach can be effective in practice on typical GPUs, we also developed a custom CUDA implementation for the ingredients of WCC. On top of that, having a custom implementation allows for several opportunities to further speed up the process, and keep the memory bandwidth low in certain common scenarios, as we detail below. 

The key ingredients of our custom implementation are as follows:
\begin{enumerate}
    \item An in-place implementation of the forward and inverse Haar transforms, using a single memory read and write for all levels. This results in an inference time for the transforms, which is comparable to double the one of average pooling.
    \item A custom gather and scatter kernels that use a single index list for all channels. 
    \item A kernel for a fused depthwise convolution and Haar transform. Since our framework rely on the idea of separable depthwise convolutions, it is natural to fuse together consecutive separable operations like depthwise convolution and the Haar transform that typically follows it. This saves memory read and write, as well as several computations that are joint for both operations (because both are separable).
\end{enumerate}

To demonstrate the effectiveness of our implementation, we compare the inference time of an inverted bottleneck residual block used in modern architectures like MobileNets \cite{sandler2018mobilenetv2, howard2019searching}, ConvNext \cite{liu2022convnext}, and EfficientNets \cite{tan2019efficientnet,tan2021efficientnetv2}. The inverted residual block that we test for timing purposes reads
\begin{equation}\label{eq:inverted}
    \bfx^{(l+1)} = \bfx^{(l)} + \bfK^{l_3}_{1\times1}\left(\bfK^{l_2}_{dw}\left(\bfK^{l_1}_{1\times1}\bfx^{(l)}\right)\right),
\end{equation}
where $\bfK^{l_3}_{1\times1}, \bfK^{l_1}_{1\times1}$ are $1\times 1$ convolutions and $\bfK^{l_2}_{dw}$ which is typically applied on a much larger channel dimension than that of $\bfx^{(l)}$ (the factor between the channel sizes if often called \emph{expansion}). We note that for the purpose of timings, we omit the non-linear activations which are typically fused into the convolution kernels are are applied at negligible cost, if simple. The timings were obtained using PyTorch, that is bounded to CUDA kernels using the package {\tt ctypes}, and is run on an NVIDIA 1080ti GPU on an isolated Linux machine. All runs are applied using the maximal batch size possible, and are averaged over 100 trials.

\autoref{Table:timing} summarizes the results for different parameters.  It is clear, as expected, that the speedup is better for lower shrinkage rates, and when the number of channels is higher. The latter is a key theoretical aspect of the speedup - the complexity of $1\times1$ convolutions is quadratic in the number of channels, while the complexity of the WCC additional operations is linear. Hence, we expect more speedup as the number of channels grows in the future. We would like to stress that (1) our implementation can probably be further optimized and (2) server GPUs may be far from the typical prototype low-resource edge device in common scenarios. 

\textbf{Memory bandwidth and traffic}: The most complicated and optimized operation of CNNs is the dense matrix-matrix multiplication, i.e., the $1\times 1$ convolution. Typically, a tile (part of an image) from \emph{all channels} has to be read by each group of threads to compute a tile of an output channel. That is in addition to reading the relevant weights. In our work, we ease these memory reads by simply reducing the dimensions of the feature maps. The other operations in the network are separable (activations, Haar, gather/scatter, depthwise convolutions) and are of linear complexity in their memory reads. Hence, as more channels are used, the relative memory traffic using WCC compared to a standard convolution will decrease. Furthermore, considering a common inverted residual block as in \eqref{eq:inverted} with a large expansion, the peak memory lies in the input and output of $\bfK_{dw}^{l_2}$. Using our method, we may apply the Haar transforms, depthwise convolutions, and scatter/gather separately and in parts using relatively small intermediate allocated memory and write the full result in a compressed form. This way, we only store the full result in a compressed manner towards the input of $\bfK_{1\times 1}^{l_3}$, which needs to be complete before the $1\times 1$ convolution. Other scenarios for memory savings can be obtained for different architectures, devices, and scenarios. 

\begin{table}[ht]
\centering
\caption{Inference timing results.}
\setlength\tabcolsep{3pt}
\begin{tabular}{cccccccc}
\toprule
Image size & $c_{in}$ & Expansion &  Batch & Comp. rate & Standard [s] & Ours [s] & Speedup\\ 
\midrule
$96\times96$ & $512$ & 2 & 48 & 0.25 & $1.82\cdot10^{-1}$ & $1.19\cdot10^{-1}$& $\times$1.52  \\
$96\times96$ & $512$ & 2 & 48 & 0.5 & $1.82\cdot10^{-1}$ & $1.52\cdot10^{-1}$& $\times1.20$\\
$96\times96$ & $512$ & 4 & 24 & 0.25 & $1.78\cdot10^{-1}$ & $1.04\cdot10^{-1}$& $\times1.71$ \\
$96\times96$ & $512$ & 4 & 24 & 0.5 & $1.76\cdot10^{-1}$ & $1.36\cdot10^{-1}$& $\times1.29$\\
$96\times96$ & $1024$ & 2 & 12 & 0.25 & $1.38\cdot10^{-1}$ & $7.11\cdot10^{-2}$& $\times1.94$ \\
$96\times96$ & $1024$ & 2 & 12 & 0.5 & $1.38\cdot10^{-1}$ & $1.00\cdot10^{-1}$ & $\times1.38$\\
$96\times96$ & $1024$ & 4 & 6 & 0.25 & $1.36\cdot10^{-1}$ & $6.36\cdot10^{-2}$ & $\times2.13$\\
$96\times96$ & $1024$ & 4 & 6 & 0.5 & $1.36\cdot10^{-1}$ & $9.20\cdot10^{-2}$ & $\times1.47$\\
%$128\times128$ & $256$ & 2 & 64 & 0.25 & $1.69\cdot10^{-1}$ & $1.38\cdot10^{-1}$& $\times$1.22 \\
$128\times128$ & $128$ & 4 & 80 & 0.25 & $1.67\cdot10^{-1}$ & $1.39\cdot10^{-1}$ & $\times$1.20\\ 
$128\times128$ & $128$ & 6 & 64 & 0.25 & $1.99\cdot10^{-1}$ & $1.56\cdot10^{-1}$ & $\times$1.28 \\  
$128\times128$ & $256$ & 4 & 32 & 0.125 & $1.65\cdot10^{-1}$ & $0.91\cdot10^{-2}$& $\times$1.81\\
$128\times128$ & $256$ & 4 & 32 & 0.25 & $1.66\cdot10^{-1}$ & $1.08\cdot10^{-1}$& $\times$1.53\\
$128\times128$ & $256$ & 4 & 32 & 0.5 & $1.66\cdot10^{-1}$ & $1.31\cdot10^{-1}$& $\times$1.27 \\
$256\times256$ & $256$ & 4 & 8 & 0.125 & $1.72\cdot10^{-1}$ & $9.27\cdot10^{-2}$& $\times1.85$\\
$256\times256$ & $256$ & 4 & 8 & 0.25 & $1.72\cdot10^{-1}$ & $1.09\cdot10^{-1}$ & $\times1.57$\\
$256\times256$ & $256$ & 4 & 8 & 0.5 & $1.69\cdot10^{-1}$ & $1.31\cdot10^{-1}$ & $\times1.29$\\
$256\times256$ & $256$ & 8 & 4 & 0.125 & $1.68\cdot10^{-1}$ & $8.45\cdot10^{-2}$& $\times1.99$\\
$256\times256$ & $256$ & 8 & 4 & 0.25 & $1.68\cdot10^{-1}$ & $1.00\cdot10^{-1}$& $\times1.68$\\
$256\times256$ & $256$ & 8 & 4 & 0.5 & $1.68\cdot10^{-1}$ & $1.21\cdot10^{-1}$& $\times1.39$\\
$512\times512$ & $128$ & 6 & 4 & 0.25 & $2.13\cdot10^{-1}$ & $1.51\cdot10^{-1}$ & $\times1.41$\\
$512\times512$ & $128$ & 6 & 2 & 0.5 & $1.04\cdot10^{-1}$ & $9.21\cdot10^{-2}$& $\times1.12$\\
$512\times512$ & $256$ & 4 & 2 & 0.125 & $1.74\cdot10^{-1}$ & $9.85\cdot10^{-2}$& $\times1.77$\\
$512\times512$ & $256$ & 4 & 2 & 0.25 & $1.73\cdot10^{-1}$ & $1.14\cdot10^{-1}$ & $\times1.52$\\
$512\times512$ & $256$ & 4 & 2 & 0.5 & $1.72\cdot10^{-1}$ & $1.43\cdot10^{-1}$ & $\times1.20$\\
%\midrule
%\midrule
\bottomrule
\end{tabular}
\label{Table:timing}
\end{table}

\end{document}